\documentclass[10pt,twocolumn,letterpaper]{article}
\usepackage{preprint}
\usepackage{multirow}
\usepackage{array}
\usepackage{tabularx}
\usepackage{makecell}
\usepackage{bm}
\usepackage{colortbl}
\usepackage{tikz}
\usetikzlibrary{arrows.meta,positioning,calc}
\usepackage[final]{microtype}   %

\definecolor{oursrow}{HTML}{FCECEE}
\definecolor{bestc}{HTML}{B52E42}
\definecolor{bestcell}{HTML}{F8DDE1}
\definecolor{secondc}{HTML}{276E76}
\definecolor{secondcell}{HTML}{E5F2EF}
\definecolor{headerblue}{HTML}{E4ECF6}
\definecolor{headermint}{HTML}{E3F1ED}
\definecolor{headerlilac}{HTML}{EDE8F5}
\definecolor{headersage}{HTML}{EAF0E3}
\definecolor{tableink}{HTML}{000000}
\definecolor{baselinebg}{HTML}{F0F4FA}
\definecolor{ablationbg}{HTML}{EEF6F2}

\newcommand{\best}[1]{\cellcolor{bestcell}\textcolor{bestc}{\textbf{#1}}}
\newcommand{\second}[1]{\cellcolor{secondcell}\textcolor{secondc}{\underline{#1}}}
\newcommand{\ours}{\rowcolor{oursrow}}
\newcommand{\oursname}[1]{\textcolor{bestc}{#1}}
\newcommand{\metricheader}[1]{\cellcolor{headerblue}\textcolor{tableink}{#1}}
\newcommand{\auxheader}[1]{\cellcolor{headermint}\textcolor{tableink}{#1}}
\newcommand{\altmetricheader}[1]{\cellcolor{headerlilac}\textcolor{tableink}{#1}}
\newcommand{\costheader}[1]{\cellcolor{headersage}\textcolor{tableink}{#1}}
\newcommand{\baselineband}{\rowcolor{baselinebg}}

\definecolor{metricblue}{HTML}{5A98C6}
\definecolor{metricpurple}{HTML}{9A7BBE}
\definecolor{metricteal}{HTML}{49A48C}
\definecolor{metricgold}{HTML}{D5AB5B}
\definecolor{metricgray}{HTML}{A9AFB8}
\definecolor{headerwarm}{HTML}{F6EBD4}
\newcommand{\warmheader}[1]{\cellcolor{headerwarm}\textcolor{tableink}{#1}}
\newcommand{\heatcell}[3]{\cellcolor{#1!#2!white}#3}
\newcommand{\methodchip}[1]{\makebox[6pt][l]{\textcolor[HTML]{#1}{\rule{2.5pt}{6pt}}}}

\definecolor{tablesd}{HTML}{4E7184}
\definecolor{tableneutral}{HTML}{F0F2F5}
\definecolor{tablebasebg}{HTML}{F5F7FA}
\definecolor{tablefinalbg}{HTML}{FDF2F4}
\definecolor{tablestablebg}{HTML}{EDF7F3}
\newcommand{\tabunc}[1]{\mathbin{\textcolor{tablesd}{\pm}}{\color{tablesd}#1}}
\newcommand{\tabbest}[1]{\textcolor{bestc}{\ifmmode\mathbf{#1}\else\textbf{#1}\fi}}
\newcommand{\tabstable}[1]{\textcolor{secondc}{\ifmmode\mathbf{#1}\else\textbf{#1}\fi}}
\newcommand{\neutralheader}[1]{\cellcolor{tableneutral}\textcolor{tableink}{#1}}
\newcommand{\tabbaseline}{\rowcolor{tablebasebg}}
\newcommand{\tabfinal}{\rowcolor{tablefinalbg}}
\newcommand{\tabstablerow}{\rowcolor{tablestablebg}}

\definecolor{captionkey}{HTML}{D7191C}
\newcommand{\capkey}[1]{\textcolor{captionkey}{\textbf{#1}}}

\newcommand{\mapfive}{mAP@5$^\circ$\xspace}

\newcommand{\Fft}{\operatorname{FFT}}
\newcommand{\Ifft}{\operatorname{IFFT}}
\newcommand{\LN}{\operatorname{LN}}
\newcommand{\MLP}{\operatorname{MLP}}
\newcommand{\Flip}{\operatorname{Flip}}
\newcommand{\Interp}{\operatorname{Interp}}
\newcommand{\BMM}{\operatorname{BMM}_N}
\newcommand{\Softmax}{\operatorname{Softmax}}
\newcommand{\dC}{\Delta\mathbf{C}}

\usepackage{algorithm}
\usepackage[noend]{algpseudocode}
\renewcommand{\thealgorithm}{S\arabic{algorithm}}
\definecolor{alggeometry}{HTML}{0B6FA8}
\definecolor{algordering}{HTML}{2A2FA5}
\definecolor{algspectral}{HTML}{C5415D}
\newcommand{\alggeo}[1]{\textcolor{alggeometry}{#1}}
\newcommand{\algord}[1]{\textcolor{algordering}{#1}}
\newcommand{\algspec}[1]{\textcolor{algspectral}{#1}}
\algrenewcommand\algorithmicrequire{\textbf{Input:}}
\algrenewcommand\algorithmicensure{\textbf{Output:}}
\algrenewcommand\alglinenumber[1]{\normalfont\scriptsize #1:}
\algrenewcommand\algorithmicindent{1.15em}

\definecolor{suppalgblue}{HTML}{1F77B4}
\definecolor{suppalgsky}{HTML}{87CEEB}
\colorlet{suppalgbody}{suppalgsky!16!white}
\newsavebox{\suppalgcontentbox}
\makeatletter
\newcommand{\fs@suppalgorithmblue}{%
  \def\@fs@cfont{\bfseries}%
  \let\@fs@capt\floatc@ruled
  \def\@fs@pre{}%
  \def\@fs@mid{\nointerlineskip}%
  \def\@fs@post{}%
  \let\@fs@iftopcapt\iftrue
}
\makeatother
\DeclareCaptionFormat{suppalgorithmblue}{%
  \setlength{\fboxsep}{6pt}%
  \colorbox{suppalgblue}{%
    \parbox{\dimexpr\linewidth-2\fboxsep\relax}{%
      \color{white}\strut #1#2#3}}%
}
\DeclareCaptionStyle{suppalgorithmblue}{%
  format=suppalgorithmblue,labelfont={bf,color=white},textfont={color=white},%
  labelsep=space,justification=raggedright,singlelinecheck=false,skip=0pt%
}
\newcommand{\suppalgorithmblue}{%
  \floatstyle{suppalgorithmblue}\restylefloat{algorithm}%
}
\newenvironment{suppalgorithmcontent}{%
  \par\noindent
  \setlength{\fboxsep}{6pt}%
  \begin{lrbox}{\suppalgcontentbox}%
  \begin{minipage}{\dimexpr\linewidth-2\fboxsep\relax}%
  \color{black}%
}{%
  \end{minipage}%
  \end{lrbox}%
  \colorbox{suppalgbody}{\usebox{\suppalgcontentbox}}\par
}

\renewcommand{\thealgorithm}{\arabic{algorithm}}
\usepackage{tcolorbox}

\definecolor{ccCluster}{HTML}{A0127A}
\definecolor{ccGateFill}{HTML}{F8CBCA}
\definecolor{ccGateLine}{HTML}{E36E70}
\definecolor{ccMambaFill}{HTML}{D9B3E4}
\definecolor{ccMambaLine}{HTML}{7030A0}
\definecolor{ccFlipFill}{HTML}{F1D68E}
\definecolor{ccFlipLine}{HTML}{BA8800}
\definecolor{ccLNFill}{HTML}{ACD78E}
\definecolor{ccLNLine}{HTML}{82B366}
\definecolor{ccInlier}{HTML}{0F9D3A}
\definecolor{ccOutlier}{HTML}{E8402A}
\definecolor{ccStrandOut}{HTML}{E8532B}

\newtcbox{\ccchipbox}[1][]{on line, boxsep=0pt, left=1.0pt, right=1.0pt, top=0.6pt, bottom=0.6pt,
  arc=1.6pt, boxrule=0.65pt, #1}
\newcommand{\ccchip}[4][black]{\text{\ccchipbox[colback=#2,colframe=#3,coltext=#1]{\ensuremath{#4}}}}

\newcommand{\kGeo}[1]{{\color{alggeometry}#1}}
\newcommand{\kOrd}[1]{{\color{algordering}#1}}
\newcommand{\kClu}[1]{{\color{ccCluster}#1}}

\newcommand{\cGeo}[1]{\ccchip[alggeometry!62!black]{alggeometry!24}{alggeometry}{#1}}
\newcommand{\cOrd}[1]{\ccchip[algordering!70!black]{algordering!20}{algordering}{#1}}
\newcommand{\cClu}[1]{\ccchip[ccCluster!62!black]{ccCluster!18}{ccCluster}{#1}}
\newcommand{\cGate}[1]{\ccchip[ccGateLine!45!black]{ccGateFill}{ccGateLine}{#1}}

\newcommand{\cMamba}[1]{\ccchip{ccMambaFill}{ccMambaLine}{#1}}
\newcommand{\cFlip}[1]{\ccchip{ccFlipFill}{ccFlipLine}{#1}}
\newcommand{\cLN}[1]{\ccchip{ccLNFill}{ccLNLine}{#1}}

\newcommand{\cclegend}[1]{\,\textcolor[HTML]{#1}{\rule[0.05ex]{0.62em}{0.62em}}}

\newcommand{\lgLowpass}{\cclegend{168E83}}
\newcommand{\lgMLP}{\cclegend{8961B6}}
\newcommand{\lgLearned}{\cclegend{C5415D}}

\newcommand{\inl}[1]{\textcolor{ccInlier}{#1}}
\newcommand{\outl}[1]{\textcolor{ccOutlier}{#1}}
\newcommand{\strandout}[1]{\textcolor{ccStrandOut}{#1}}

\usepackage{amsthm}
\newtheorem{proposition}{Proposition}
\newtheorem{corollary}{Corollary}

\definecolor{preprintblue}{RGB}{37,82,125}
\usepackage[breaklinks,colorlinks,allcolors=preprintblue]{hyperref}
\usepackage[capitalize]{cleveref}
\crefname{section}{Sec.}{Secs.}
\Crefname{section}{Section}{Sections}
\crefname{table}{Tab.}{Tabs.}
\Crefname{table}{Table}{Tables}
\hypersetup{
  pdftitle={Smooth Inliers, Erratic Outliers: Frequency-Aware Mamba for Two-View Correspondence Learning},
  pdfauthor={Zhihua Wang, Yifeng Gao, Yunhao Feng, Rui Pan, Zhouyuan Ma, Baiheng Zhao, Zixing Chen, Yanping Li, Yizhang Liu, Xingjun Ma},
  pdfsubject={Preprint},
  pdfkeywords={correspondence learning, Mamba, frequency-domain gating, geometric vision},
  bookmarksnumbered=true
}
\title{Smooth Inliers, Erratic Outliers: Frequency-Aware Mamba for Two-View Correspondence Learning}
\author{%
  \textbf{\mbox{Zhihua Wang}}\textsuperscript{1,2}\thanks{This work was completed during Zhihua Wang's internship at Fudan University.}\quad
  \textbf{\mbox{Yifeng Gao}}\textsuperscript{1}\quad
  \textbf{\mbox{Yunhao Feng}}\textsuperscript{3}\quad
  \textbf{\mbox{Rui Pan}}\textsuperscript{1}\quad
  \textbf{\mbox{Zhouyuan Ma}}\textsuperscript{1}\quad
  \textbf{\mbox{Baiheng Zhao}}\textsuperscript{1}\quad
  \textbf{\mbox{Zixing Chen}}\textsuperscript{1}\quad
  \textbf{\mbox{Yanping Li}}\textsuperscript{4}\quad
  \textbf{\mbox{Yizhang Liu}}\textsuperscript{5}\quad
  \textbf{\mbox{Xingjun Ma}}\textsuperscript{1,6}\thanks{Correspondence: \href{mailto:xingjunma@fudan.edu.cn}{\texttt{xingjunma@fudan.edu.cn}}}\\[7pt]
  {\normalfont\small
  \mbox{\textsuperscript{1}Fudan University}\quad
  \mbox{\textsuperscript{2}University of Shanghai for Science and Technology}\quad
  \mbox{\textsuperscript{3}Ant Group}\quad
  \mbox{\textsuperscript{4}Shanghai Jiao Tong University}\quad
  \mbox{\textsuperscript{5}Fuzhou University}\quad
  \mbox{\textsuperscript{6}Shanghai Innovation Institute}}%
}
\date{}
\begin{document}
\maketitle
\begin{abstract}
When shown matches between two photographs of the same cathedral, people can tell at a glance which ones are wrong: a window matched to its look-alike jumps against its neighbors, while correct matches shift together.
Two-view correspondence learning, also called correspondence pruning, asks a network to make this judgment for thousands of candidate matches, most of which are wrong.
Mamba, a fast sequence model, now drives strong pruners, but it reads the matches as a plain sequence, and whether its scan can capture such agreement is unclear.
We propose \textbf{SFMambaNet}, which scans the matches cluster by cluster, so that correct matches vary slowly along the sequence while wrong ones stay erratic, and filters the scanned features with learnable frequency-domain gates between its forward and backward passes.
On camera pose estimation, outlier removal and homography estimation, we find that \textbf{SFMambaNet} outperforms spatial-domain Mamba baselines with about two million parameters.
Contrary to the usual expectation that such gains come from extra capacity, we find that a parameter-free low-pass filter beats point-wise layers as large as the gates.
The learned gates turn low-pass on their own and make the scan far less sensitive to how the matches are shuffled.
Together, these results show that frequency is a natural language for describing agreement between correspondences, with implications for other sequence models that must read unordered sets, such as point clouds and graphs.
\end{abstract}

\section{Introduction}
\label{sec:intro}

\begin{figure}[t]
  \centering
  \includegraphics[width=\linewidth]{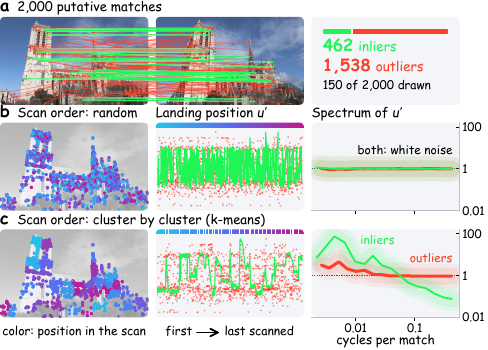}
  \caption{\textbf{Smooth inliers, erratic outliers.}
  \textbf{(a)}~Putative matches.
  \textbf{(b)}~In a random scan order, where they land in the second image is noise for both classes.
  \textbf{(c)}~Cluster by cluster, \inl{inliers} land in smooth, low-frequency steps; \outl{outliers} stay erratic (Appendix~K).}
  \label{fig:teaser}
\end{figure}

\begin{figure*}[t]
  \centering
  \includegraphics[width=\textwidth]{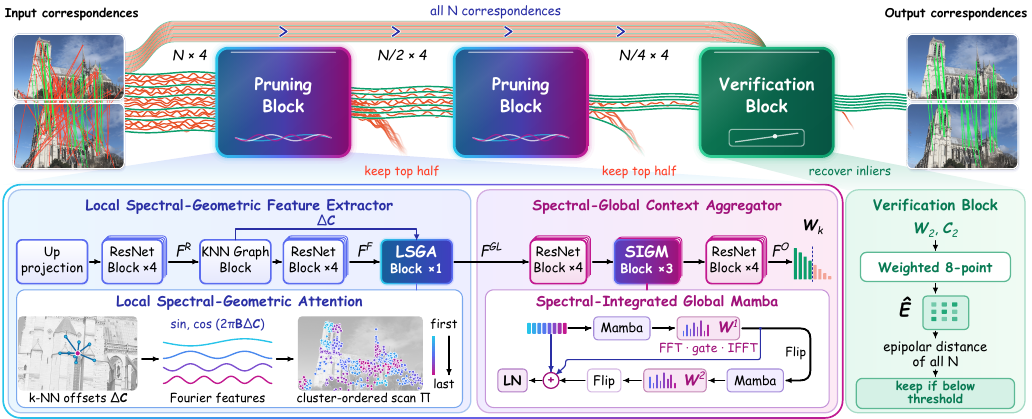}
  \caption{\textbf{Architecture of SFMambaNet.} \capkey{Top:} the $N\times4$ putative correspondences, shown as strands (\inl{green}: inliers, \strandout{orange}: outliers), pass through two pruning blocks, each of which keeps the more reliable half, and a verification block that estimates the essential matrix and re-checks all correspondences. \capkey{Bottom:} a pruning block consists of a Local Spectral-Geometric Feature Extractor, built around the LSGA block (\cref{sec:lsga}), and a Spectral-Global Context Aggregator, built around three SIGM blocks (\cref{sec:sigm}).}
  \label{fig:arch}
\end{figure*}

Human vision is remarkably good at deciding what belongs together.
Elements that move as one are seen as one, a grouping principle that Gestalt psychology calls common fate~\cite{Wertheimer1923Gestalt,Wagemans2012Gestalt}, and an early computational theory of stereo vision rests on a related observation: disparity varies smoothly almost everywhere~\cite{Marr1976Stereo}.
We rarely judge a single pairing of points in isolation; we judge it by how well it agrees with its surroundings.
For example, when we compare two photographs of the same cathedral taken from different spots, we can tell at a glance which details correspond: true pairs shift together with their neighbors, whereas a window paired with a look-alike window jumps against them.

Computer vision faces the same question at a much larger scale.
Correspondences between two images are the starting point of visual localization~\cite{Sattler2018Benchmarking6DoF,Sarlin2019CoarseToFine}, image registration~\cite{Ma2015RobustFeatureMatchingLLT}, SLAM~\cite{MurArtal2015ORBSLAM}, and structure from motion~\cite{Schonberger2016SfMRevisited,Goesele2007MultiViewStereo}, and they are commonly obtained by matching SIFT~\cite{Lowe2004SIFT} or SuperPoint~\cite{DeTone2018SuperPoint} descriptors to their nearest neighbors.
Under wide baselines, repeated textures, or illumination changes, most of the resulting putative correspondences are wrong (\cref{fig:teaser}a), and hypothesize-and-verify estimators such as RANSAC~\cite{Fischler1981RANSAC,Chum2005PROSAC,Torr2000MLESAC,Barath2019MAGSAC} slow down sharply as the inlier ratio drops~\cite{Raguram2012USAC}.
\emph{Correspondence pruning}\footnote{Also called two-view correspondence learning; we use both terms interchangeably.} turns the human judgment into a learning problem: since LFGC~\cite{Yi2018GoodCorrespondences}, a network looks at the whole set at once, classifies every correspondence as an inlier or an outlier, and passes the result on to a differentiable eight-point solver that recovers the relative pose between the two cameras~\cite{Zhang2019OrderAwareNetwork,Sun2020ACNe,Zhong2021TNet,Zhao2021ProgressiveCorrespondencePruning}.

\suppressfloats[t]
\begin{figure}[t]
\centering
\includegraphics[width=\linewidth]{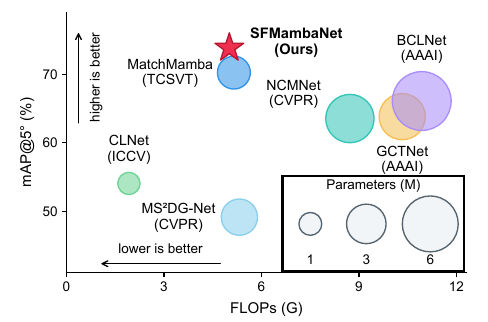}
\caption{\textbf{Accuracy--cost trade-off} on YFCC100M Unknown with SIFT and no RANSAC. SFMambaNet (\textcolor[HTML]{EE2F4F}{red star}) is the most accurate at a slightly lower cost than MatchMamba; circle area is proportional to parameters. Baseline values follow~\cite{Wu2025MatchMamba}; exact costs are in Appendix~J.}
\label{fig:efficiency_costs}
\end{figure}

Over the past years, pruners have become steadily better at measuring agreement, both with neighbors and with a single epipolar geometry.
Graph networks mine local agreement over $k$-nearest-neighbor graphs~\cite{Zhao2019NMNet,Zhao2021ProgressiveCorrespondencePruning,Dai2022MS2DGNet,Li2023UMatch,Liu2023ProgressiveNeighborConsistencyMining,Dai2024MGNet}, Transformers capture global agreement at quadratic cost~\cite{LiaoVol38VSFormer,LiuVol159TransMatch,Miao2024BCLNet,Dai2026LeCoT}, and selective state space models such as Mamba~\cite{Gu2023Mamba,Dao2024Mamba2} now bring linear-time global context to pruning~\cite{Wu2025MatchMamba,miao2025mambamatch,Fang2026CorrMamba}.
The intuition that correct matches move together has also been built in directly: classical methods enforce the coherence of correct motion~\cite{Ma2014VectorFieldConsensus,Bian2017GMS,Ma2019LocalityPreservingMatching,Fan2023SmoothnessDrivenConsensus,Xia2025GGSLC}, and motion-field networks turn it into explicit smoothness priors~\cite{Liu2021LMCNet,Zhang2023ConvMatch,Zhang2024DeMatch}.
For Mamba-based pruners, two questions remain open.
The first concerns order: Mamba was designed for sequences, whereas correspondences form a set, so the scan reads them in an arbitrary order and hands every feature, including the erratic response of an outlier, from one step to the next.
The second concerns representation: consistency is inferred from coarse four-dimensional coordinates, and networks fed with such inputs favor smooth functions~\cite{Rahaman2019SpectralBias,Tancik2020FourierFeatures}, which blurs the small offsets that separate an inlier from a nearby outlier.

We close these gaps by looking at agreement through frequency.\footnote{Throughout, frequency is the index of the discrete Fourier transform along a scan order; low frequencies change slowly from one correspondence to the next.}
Frequency needs an order: in a random order, every correspondence looks like noise (\cref{fig:teaser}b), but scanned cluster by cluster, inliers fall into smooth, low-frequency steps while outliers stay erratic (\cref{fig:teaser}c); on this pair, we find that the lowest tenth of the frequencies holds 84\% of the inliers' energy but only 15\% of the outliers'.
Building on this observation, we introduce \textbf{SFMambaNet} (\cref{fig:arch}), which brings frequency into the scan at both scales.
Locally, its \emph{Local Spectral-Geometric Attention} (LSGA) block expands relative coordinates into Fourier features before attending over neighbors, and groups the correspondences into learned clusters whose order defines the scan.
Globally, its \emph{Spectral-Integrated Global Mamba} (SIGM) block scans along this cluster order and places learnable frequency-domain gates between the forward and the backward scan, so that the second scan receives filtered features.
Along the cluster order, frequency gains a clear meaning: slow variation is agreement within a cluster, while the arbitrary order inside a cluster only adds deviations with little low-frequency energy (\cref{sec:sigm}).
\Cref{fig:efficiency_costs} previews the outcome: higher accuracy than every compared pruner at a slightly lower cost than MatchMamba.
Through benchmarks and controlled experiments, we then make the following contributions and findings:
\begin{enumerate}[label=\arabic*.,noitemsep,leftmargin=*,topsep=0em]
\item We show that frequency modeling fits inside a Mamba pruner at little cost: Fourier-feature attention (LSGA) and spectral gates between cluster-ordered scans (SIGM) add only $\mathcal{O}(DN\log N)$ operations per scan (\cref{sec:method}).
\item We show that SFMambaNet outperforms the compared pruners, including MatchMamba~\cite{Wu2025MatchMamba}, on relative pose estimation (YFCC100M and SUN3D, SIFT and SuperPoint), outlier removal, and homography estimation, with only 2.01M parameters and 5.0 GFLOPs (\cref{sec:experiments}).
\item We find that gains come from filtering, not capacity: a parameter-free low-pass filter beats equally large point-wise layers, and learned gates turn low-pass (\cref{sec:analysis}).
\item We find that these gates make the scan far less sensitive to shuffling, and that Fourier features help through frequency, not dimension (\cref{sec:analysis}).
\end{enumerate}

\section{Method}
\label{sec:method}

\begin{figure*}[t]
  \centering
  \includegraphics[width=\textwidth]{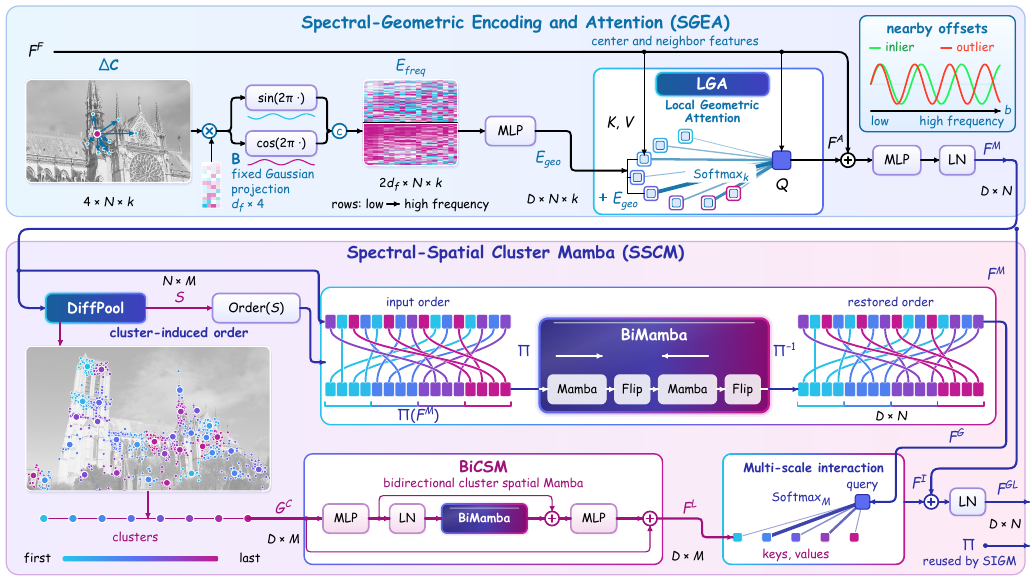}
  \caption{\textbf{LSGA block.} \capkey{Top:} spectral-geometric encoding and attention (SGEA) expands the relative coordinates $\kGeo{\Delta C}$ with a fixed Gaussian projection and sine/cosine features (\cref{eq:sge}) and injects them into a local geometric attention (LGA, \cref{eq:qkv,eq:lga}). \capkey{Bottom:} the spectral-spatial cluster Mamba (SSCM) scans the point features with a bidirectional Mamba along the cluster-induced order and refines the DiffPool clusters with a bidirectional cluster spatial Mamba (BiCSM); a multi-scale interaction lets every correspondence query the cluster consensus (\cref{eq:msi}). Colors mark the scan position in $\kOrd{\Pi}$; the clusters shown are the k-means blocks of \cref{fig:teaser}c.}
  \label{fig:lsga}
\end{figure*}

\subsection{Problem formulation and overview}
\label{sec:overview}

Given two images, we first detect keypoints with SIFT~\cite{Lowe2004SIFT} or SuperPoint~\cite{DeTone2018SuperPoint} and match their descriptors by nearest-neighbor search, obtaining the putative set
\begin{equation}
  C=\{c_1,c_2,\dots,c_N\}\in\mathbb{R}^{N\times4},\quad c_i=(u_i,v_i,u'_i,v'_i),
  \label{eq:input}
\end{equation}
where $(u_i,v_i)$ and $(u'_i,v'_i)$ are normalized keypoint coordinates in the two images.
The goal is to identify the correct correspondences in $C$.
Following the iterative pruning paradigm~\cite{Zhao2021ProgressiveCorrespondencePruning,Dai2022MS2DGNet}, SFMambaNet (\cref{fig:arch}, top) applies two pruning blocks and a verification block:
\begin{equation}
\begin{gathered}
  (\mathcal{W}_1,C_1)=f_{\theta_1}(C),\quad (\mathcal{W}_2,C_2)=f_{\theta_2}(\mathcal{W}_1,C_1),\\
  \hat{\mathbf{E}}=g(\mathcal{W}_2,C_2),\quad \mathcal{D}=h(\hat{\mathbf{E}},C),
\end{gathered}
\label{eq:pipeline}
\end{equation}
where $(\mathcal{W}_k,C_k)$ are the predicted inlier weights and the retained subset after the $k$-th block, $g$ is the differentiable weighted eight-point algorithm, and $h$ computes the epipolar distances of \emph{all} correspondences to $\hat{\mathbf{E}}$.
Correspondences whose distance falls below a threshold are kept, which recovers inliers removed during pruning.

Inside a pruning block (\cref{fig:arch}, bottom), $C$ is embedded into $D$ channels and refined by ResNet blocks into $F^R$.
A local graph connects every correspondence with its $k$ nearest neighbors under the sparse semantic distance of~\cite{Dai2022MS2DGNet}; edge features $[f^R_i,\,f^R_i-f^R_{ij}]$ are fused by an MLP and max pooling into the graph feature $F^F\in\mathbb{R}^{D\times N\times1}$.
Max pooling, however, keeps the strongest activations and suppresses the small perturbations that separate inliers from nearby outliers.
LSGA restores them in the frequency domain.

\subsection{Local Spectral-Geometric Attention}
\label{sec:lsga}

LSGA (\cref{fig:lsga}) consists of a spectral-geometric encoding and attention (SGEA) step, which refines every correspondence with its neighborhood, and a spectral-spatial cluster Mamba (SSCM), which adds region-level consensus.

\paragraph{Spectral-geometric encoding.}
For every $c_i$ we gather the coordinates of its $k$ graph neighbors into $C_{\mathcal{N}}\in\mathbb{R}^{4\times N\times k}$ and form the relative coordinates $\cGeo{\dC}=C_{\mathcal{N}}-c_i$.
Instead of feeding these low-dimensional offsets to an MLP, which is biased towards smooth functions of its input~\cite{Tancik2020FourierFeatures}, we expand them into a Fourier-feature spectrum with a fixed random Gaussian projection $\cGeo{\mathbf{B}}\in\mathbb{R}^{d_f\times4}$ shared by all neighbors:
\begin{equation}
  \kGeo{E_{\mathit{freq}}}=\big[\sin(2\pi\kGeo{\mathbf{B}\dC});\,\cos(2\pi\kGeo{\mathbf{B}\dC})\big]\in\mathbb{R}^{2d_f\times N\times k}.
  \label{eq:sge}
\end{equation}
In this spectrum, small fluctuations of the relative motion produce large changes of the encoding, so geometrically consistent neighbors become separable from erratic ones.

\paragraph{Local geometric attention.}
The encoding is mapped to a geometric embedding $\cGeo{E_{\mathit{geo}}}=\MLP(\cGeo{E_{\mathit{freq}}})\in\mathbb{R}^{D\times N\times k}$ and injected into an attention over $k$ neighbors, whose features $F^F_{\mathcal{N}}\in\mathbb{R}^{D\times N\times k}$ come from $F^F$:
\begin{equation}
\begin{gathered}
  Q=W_QF^F,\quad K=W_K\big(F^F_{\mathcal{N}}+\kGeo{E_{\mathit{geo}}}\big),\\
  V=W_V\big(F^F_{\mathcal{N}}+\kGeo{E_{\mathit{geo}}}\big),
\end{gathered}
\label{eq:qkv}
\end{equation}
\begin{equation}
\begin{gathered}
  \mathcal{A}=\Softmax_k\!\Big(\tfrac{1}{\sqrt{D}}\BMM(Q^{\top},K)\Big),\\
  F^A=\BMM(V,\mathcal{A}^{\top}),
\end{gathered}
\label{eq:lga}
\end{equation}
where $\BMM$ multiplies the matrices of every correspondence separately.
Each correspondence thus weights its neighbors by both feature similarity and spectral-geometric consistency, at $\mathcal{O}(NkD)$ instead of $\mathcal{O}(N^2D)$ cost.
A residual fusion gives the refined features $F^M=\LN\big(\MLP(F^F+F^A)\big)$; Appendix~B details the tensor operations.

\paragraph{Spectral-spatial cluster Mamba.}
Point-wise features alone do not capture region-level consensus.
We softly assign correspondences to $M\ll N$ clusters with differentiable pooling~\cite{Ying2018DiffPool} and let the assignment define a scan order:
\begin{equation}
\begin{aligned}
  \kClu{G^C},\kClu{S}&=\operatorname{DiffPool}(F^M),\quad \kOrd{\Pi}=\operatorname{Order}(\kClu{S}),\\
  F^G&=\kOrd{\Pi^{-1}}\big(\operatorname{BiMamba}(\kOrd{\Pi}(F^M))\big),
\end{aligned}
\label{eq:order}
\end{equation}
where $\cClu{S}\in\mathbb{R}^{N\times M}$ is the soft assignment, $\operatorname{Order}$ groups the correspondences by the cluster slot to which they are most strongly assigned, and $\cOrd{\Pi^{-1}}$ restores the original order.
The bidirectional Mamba therefore scans a region-aware sequence instead of an arbitrary input order.
In parallel, the cluster features $\cClu{G^C}\in\mathbb{R}^{D\times M}$ are refined by a bidirectional cluster spatial Mamba (BiCSM): following~\cite{Wu2025MatchMamba}, it transposes the spatial and channel axes, shares a perceptron across clusters, and lets the selection mechanism of Mamba keep the spatially consistent ones, yielding $\cClu{F^L}\in\mathbb{R}^{D\times M}$.
A multi-scale interaction, after the cross-scale fusion of~\cite{miao2025mambamatch}, lets each correspondence query the cluster consensus:
\begin{equation}
\begin{gathered}
  F^I=\kClu{F^L}\,\Softmax\!\Big(\tfrac{1}{\sqrt{D}}(\kClu{F^L})^{\top}F^G\Big),\\
  F^{GL}=\LN\big(F^M+F^I\big).
\end{gathered}
\label{eq:msi}
\end{equation}

\begin{figure}[t]
  \centering
  \includegraphics[width=\linewidth]{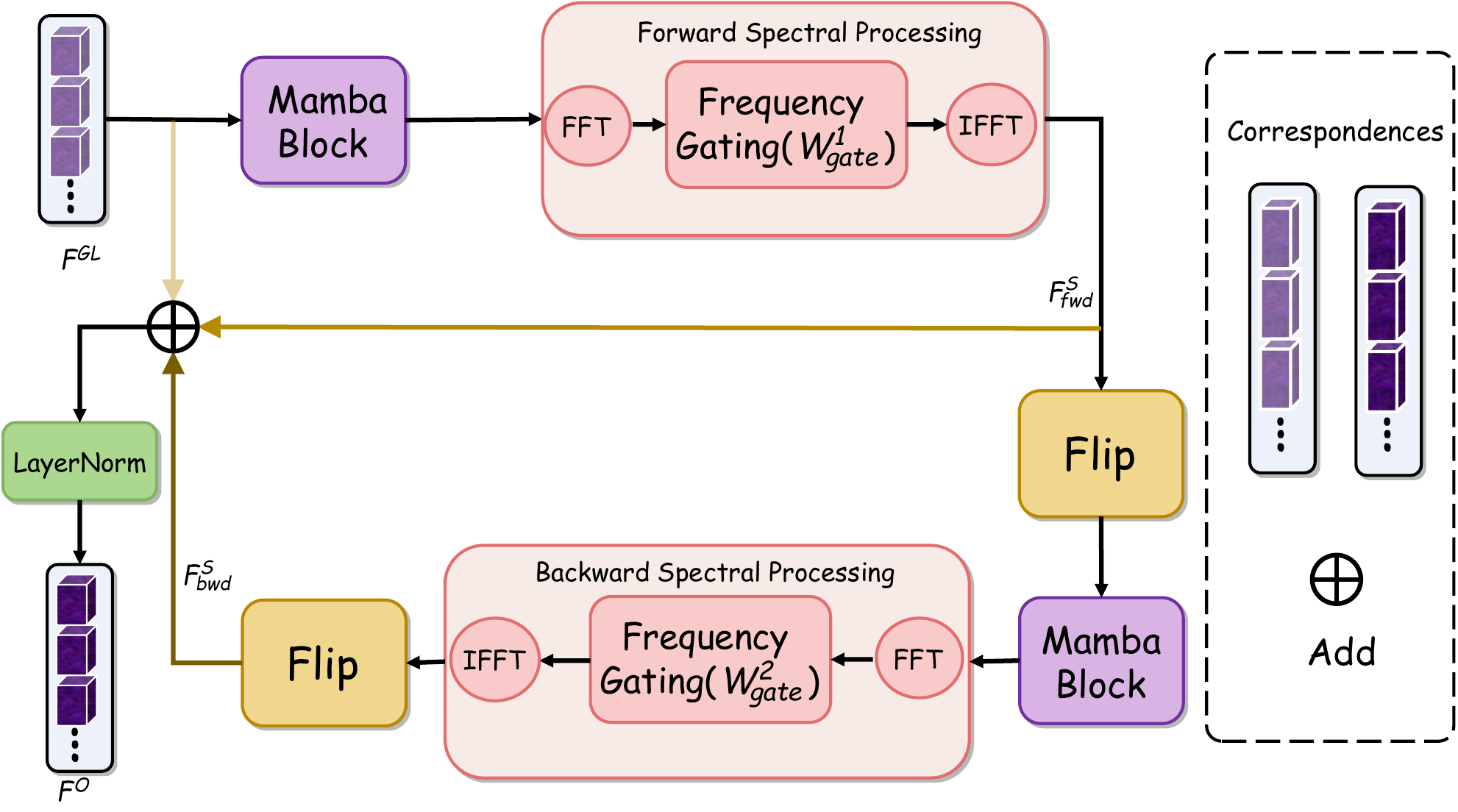}
  \caption{\textbf{SIGM block.} A forward Mamba scan is followed by a learnable frequency-domain gate; the gated sequence is flipped, scanned backwards, and gated again. Both directions are fused with the input by a residual connection and layer normalization.}
  \label{fig:sigm}
\end{figure}

\subsection{Spectral-Integrated Global Mamba}
\label{sec:sigm}

Deciding whether a correspondence is correct ultimately requires global consensus.
Bidirectional Mamba scans, such as the flip-based scan of MatchMamba~\cite{Wu2025MatchMamba}, provide it in linear time but treat the sequence purely spatially: whatever the forward scan emits, including erratic outlier responses, is accumulated again by the backward scan.
SIGM (\cref{fig:sigm}) filters the propagated features in the frequency domain in between.
\begin{algorithm}[t]
\caption{Spectral-Integrated Global Mamba block}
\label{alg:sigm}
\small
\begin{algorithmic}[1]
\Require $F^{GL}\in\mathbb{R}^{D\times N}$, order $\cOrd{\Pi}$ (\cref{eq:order}), gates $\cGate{W^1_{\mathit{gate}}},\cGate{W^2_{\mathit{gate}}}$
\Ensure $F^O\in\mathbb{R}^{D\times N}$ in the input order
\State $F^{GL}_{\mathit{ord}}\gets\cOrd{\Pi}(F^{GL})$ \Comment{\algord{\textbf{cluster order}}}
\State $F_{\mathit{fwd}}\gets\cMamba{\operatorname{Mamba}}(F^{GL}_{\mathit{ord}})$ \Comment{forward scan}
\State $F^S_{\mathit{fwd}}\gets\Ifft\big(\Fft(F_{\mathit{fwd}})\odot\Interp(\cGate{W^1_{\mathit{gate}}})\big)$ \Comment{\algspec{\textbf{gate}}}
\State $F_{\mathit{bwd}}\gets\cMamba{\operatorname{Mamba}}\big(\cFlip{\Flip}(F^S_{\mathit{fwd}})\big)$ \Comment{scan the \textbf{gated} features}\label{alg:sigm:bwd}
\State $F^S_{\mathit{bwd}}\gets\Ifft\big(\Fft(F_{\mathit{bwd}})\odot\Interp(\cGate{W^2_{\mathit{gate}}})\big)$ \Comment{\algspec{\textbf{gate}}}
\State $Y\gets\cLN{\LN}\big(F^{GL}_{\mathit{ord}}+F^S_{\mathit{fwd}}+\cFlip{\Flip}(F^S_{\mathit{bwd}})\big)$ \Comment{fuse}
\State \Return $F^O=\cOrd{\Pi^{-1}}(Y)$ \Comment{\algord{\textbf{restore the order}}}
\end{algorithmic}
\end{algorithm}
It reuses the cluster-induced permutation $\cOrd{\Pi}$ of \cref{eq:order} to order its input and proceeds as in \cref{alg:sigm}: the FFT runs along the sequence axis of every channel, $\cGate{W^1_{\mathit{gate}}}$ and $\cGate{W^2_{\mathit{gate}}}$ are learnable complex-valued gates~\cite{Rao2021GFNet,Guibas2022AFNO}, and $\Interp$ resamples them to the current sequence length, which differs between the two pruning blocks.
The backward scan consumes the \emph{gated} forward features (line~\ref{alg:sigm:bwd}); the first gate changes its input representation, and the second transforms its output.
The gates do not alter the state transition of Mamba; they control what is propagated through it.
Being complex-valued per channel and frequency, they learn both amplitude and phase.
Each FFT costs $\mathcal{O}(DN\log N)$ and each scan $\mathcal{O}(DN)$, so SIGM remains sub-quadratic in $N$.

\paragraph{What does frequency mean on an unordered set?}
A Fourier transform needs an ordered axis, and $\cOrd{\Pi}$ supplies one in which every cluster occupies a contiguous block.
The assignment network is permutation-equivariant, so shuffling the correspondences keeps each one in its cluster and the clusters in their order; only the order inside a block changes.
Such a change only moves the zero-mean deviations from the block means, and these carry little low-frequency energy: if every block has at most $L$ entries, at most a fraction $(\pi\omega_cL/N)^2$ of their energy lies below the cutoff $\omega_c$ (Appendix~D).
Low frequencies along $\Pi$ therefore describe cluster-level structure that does not depend on the input order, and a gate that favors them is largely insensitive to the order inside clusters, as \cref{sec:analysis} confirms.

\subsection{Training objective}
\label{sec:loss}

As in~\cite{Zhang2019OrderAwareNetwork,Zhao2021ProgressiveCorrespondencePruning}, we combine a binary cross-entropy loss $\mathcal{L}_{\mathit{cls}}$ on the predicted weights with a geometric loss on the estimated essential matrix $\hat{\mathbf{E}}$ of \cref{eq:pipeline},
\begin{equation}
  \mathcal{L}_{\mathit{total}}=\mathcal{L}_{\mathit{cls}}+\gamma\,\mathcal{L}_{\mathit{geo}}(\hat{\mathbf{E}},\mathbf{E}_{gt}),
  \label{eq:loss}
\end{equation}
where $\gamma$ balances the two terms and $\mathcal{L}_{\mathit{geo}}$ measures the Sampson distance of virtual correspondences (Appendix~C).

\section{Experiments}
\label{sec:experiments}

\subsection{Setup}
\label{sec:setup}

\paragraph{Datasets and metrics.}
We follow the OANet protocol~\cite{Zhang2019OrderAwareNetwork} on outdoor YFCC100M~\cite{Thomee2016YFCC100M} and indoor SUN3D~\cite{Xiao2013SUN3D}: the \emph{known} split holds test pairs from training sequences, and the \emph{unknown} split holds sequences never seen in training (Appendix~C).
For relative pose we report the mAP of rotation and translation errors at $5^\circ$ and $20^\circ$ \emph{without} RANSAC; for outlier removal, precision (P), recall (R), and F-score (F).

\paragraph{Implementation.}
Our network has 2.01M parameters and $D=128$ channels, processes $N=2000$ putative correspondences, and is trained with Adam~\cite{Kingma2014Adam} for 500k iterations on one RTX~3090; Appendix~C lists all settings.
Prior results come from MatchMamba~\cite{Wu2025MatchMamba}, which evaluates all of them under this protocol; $\dagger$ marks MatchMamba retrained by us.

\begin{table*}[t]
\centering
\caption{\textbf{Relative pose estimation} without RANSAC (mAP, \%). Left: SIFT, \mapfive on outdoor YFCC100M and indoor SUN3D. Right: SuperPoint on YFCC100M. $\dagger$: MatchMamba retrained by us (gray, not ranked); ``--'': not available. Best results are bold, second best underlined; darker cells mark higher mAP.}
\label{tab:pose}
\setlength{\tabcolsep}{5.6pt}
\small
\begingroup
\renewcommand{\best}[1]{\textcolor{bestc}{\textbf{#1}}}
\renewcommand{\second}[1]{\textcolor{secondc}{\underline{#1}}}
\begin{tabular}{@{}l cccc c cccc@{}}
\toprule
\multicolumn{1}{@{}c}{\multirow{3}{*}[-\dimexpr\aboverulesep+\belowrulesep+\cmidrulewidth\relax]{Method}} & \multicolumn{4}{c}{\metricheader{SIFT, \mapfive}} & & \multicolumn{4}{c}{\auxheader{SuperPoint, YFCC100M}} \\
\cmidrule(lr){2-5}\cmidrule(lr){7-10}
 & \multicolumn{2}{c}{\metricheader{YFCC100M}} & \multicolumn{2}{c}{\altmetricheader{SUN3D}} & & \multicolumn{2}{c}{\auxheader{Known}} & \multicolumn{2}{c}{\warmheader{Unknown}} \\
\cmidrule(lr){2-3}\cmidrule(lr){4-5}\cmidrule(lr){7-8}\cmidrule(lr){9-10}
 & \metricheader{Known} & \metricheader{Unknown} & \altmetricheader{Known} & \altmetricheader{Unknown} & & \auxheader{$5^\circ$} & \auxheader{$20^\circ$} & \warmheader{$5^\circ$} & \warmheader{$20^\circ$} \\
\midrule
\methodchip{8F9AA8}RANSAC~\cite{Fischler1981RANSAC}                          &  \heatcell{metricblue}{12}{5.81} & \heatcell{metricblue}{12}{16.88} &  \heatcell{metricpurple}{12}{4.73} &  \heatcell{metricpurple}{12}{3.26} & & \heatcell{metricteal}{13}{12.85} & \heatcell{metricteal}{12}{31.22} & \heatcell{metricgold}{12}{17.47} & \heatcell{metricgold}{12}{38.83} \\
\methodchip{8D9C68}PointNet++~\cite{Qi2017PointNetPP}                        & \heatcell{metricblue}{16}{10.49} & \heatcell{metricblue}{12}{16.48} & \heatcell{metricpurple}{24}{10.58} &  \heatcell{metricpurple}{25}{8.10} & &  \heatcell{metricgray}{8}{--}   &  \heatcell{metricgray}{8}{--}   &  \heatcell{metricgray}{8}{--}   &  \heatcell{metricgray}{8}{--}   \\
\methodchip{A48668}LFGC~\cite{Yi2018GoodCorrespondences}                     & \heatcell{metricblue}{19}{13.81} & \heatcell{metricblue}{19}{23.95} & \heatcell{metricpurple}{31}{13.78} & \heatcell{metricpurple}{34}{11.40} & & \heatcell{metricteal}{12}{12.18} & \heatcell{metricteal}{17}{34.75} & \heatcell{metricgold}{20}{24.25} & \heatcell{metricgold}{28}{52.70} \\
\methodchip{686FA7}OANet++~\cite{Zhang2019OrderAwareNetwork}                 & \heatcell{metricblue}{36}{32.57} & \heatcell{metricblue}{32}{38.95} & \heatcell{metricpurple}{46}{20.86} & \heatcell{metricpurple}{47}{16.18} & & \heatcell{metricteal}{35}{29.52} & \heatcell{metricteal}{41}{53.76} & \heatcell{metricgold}{34}{35.27} & \heatcell{metricgold}{45}{66.81} \\
\methodchip{5E9C98}ACNe~\cite{Sun2020ACNe}                                   & \heatcell{metricblue}{33}{29.17} & \heatcell{metricblue}{26}{33.06} & \heatcell{metricpurple}{41}{18.86} & \heatcell{metricpurple}{41}{14.12} & & \heatcell{metricteal}{31}{26.72} & \heatcell{metricteal}{35}{49.29} & \heatcell{metricgold}{31}{32.98} & \heatcell{metricgold}{40}{62.68} \\
\methodchip{B58E53}SuperGlue~\cite{Sarlin2020SuperGlue}                      & \heatcell{metricblue}{38}{35.00} & \heatcell{metricblue}{40}{48.12} & \heatcell{metricpurple}{49}{22.50} & \heatcell{metricpurple}{49}{17.11} & &  \heatcell{metricgray}{8}{--}   &  \heatcell{metricgray}{8}{--}   &  \heatcell{metricgray}{8}{--}   &  \heatcell{metricgray}{8}{--}   \\
\methodchip{A27292}T-Net~\cite{Zhong2021TNet}                                & \heatcell{metricblue}{46}{44.49} & \heatcell{metricblue}{43}{52.28} & \heatcell{metricpurple}{54}{24.96} & \heatcell{metricpurple}{56}{19.71} & & \heatcell{metricteal}{42}{34.97} & \heatcell{metricteal}{46}{57.50} & \heatcell{metricgold}{40}{40.65} & \heatcell{metricgold}{49}{70.36} \\
\methodchip{72A877}CLNet~\cite{Zhao2021ProgressiveCorrespondencePruning}     & \heatcell{metricblue}{41}{39.00} & \heatcell{metricblue}{45}{54.05} & \heatcell{metricpurple}{45}{20.62} & \heatcell{metricpurple}{49}{16.95} & & \heatcell{metricteal}{32}{27.56} & \heatcell{metricteal}{37}{50.82} & \heatcell{metricgold}{38}{39.19} & \heatcell{metricgold}{46}{67.37} \\
\methodchip{6198B7}MS$^2$DG-Net~\cite{Dai2022MS2DGNet}                       & \heatcell{metricblue}{41}{38.36} & \heatcell{metricblue}{40}{49.13} & \heatcell{metricpurple}{48}{22.20} & \heatcell{metricpurple}{51}{17.84} & & \heatcell{metricteal}{37}{31.15} & \heatcell{metricteal}{43}{55.16} & \heatcell{metricgold}{38}{39.19} & \heatcell{metricgold}{49}{70.36} \\
\methodchip{A57E61}PGFNet~\cite{Liu2023PGFNet}                               & \heatcell{metricblue}{46}{44.20} & \heatcell{metricblue}{44}{53.70} & \heatcell{metricpurple}{51}{23.66} & \heatcell{metricpurple}{55}{19.32} & &  \heatcell{metricgray}{8}{--}   &  \heatcell{metricgray}{8}{--}   &  \heatcell{metricgray}{8}{--}   &  \heatcell{metricgray}{8}{--}   \\
\methodchip{A589BC}U-Match~\cite{Li2023UMatch}                               & \heatcell{metricblue}{48}{46.78} & \heatcell{metricblue}{50}{60.22} & \heatcell{metricpurple}{54}{24.99} & \heatcell{metricpurple}{52}{18.13} & & \heatcell{metricteal}{42}{35.01} & \heatcell{metricteal}{45}{56.80} & \heatcell{metricgold}{44}{44.29} & \heatcell{metricgold}{50}{70.90} \\
\methodchip{2F8C8A}NCMNet~\cite{Liu2023ProgressiveNeighborConsistencyMining} & \heatcell{metricblue}{53}{52.39} & \heatcell{metricblue}{53}{63.52} & \heatcell{metricpurple}{56}{25.72} & \heatcell{metricpurple}{59}{20.82} & & \heatcell{metricteal}{47}{38.92} & \heatcell{metricteal}{51}{61.28} & \heatcell{metricgold}{49}{48.20} & \heatcell{metricgold}{55}{74.71} \\
\methodchip{C29436}GCTNet~\cite{Guo2024GraphContextTransformation}           &  \heatcell{metricgray}{8}{--}   &  \heatcell{metricgray}{8}{--}   &  \heatcell{metricgray}{8}{--}   &  \heatcell{metricgray}{8}{--}   & & \heatcell{metricteal}{46}{37.86} & \heatcell{metricteal}{50}{60.55} & \heatcell{metricgold}{48}{47.17} & \heatcell{metricgold}{55}{74.76} \\
\methodchip{7585BA}MGNet~\cite{Dai2024MGNet}                                 & \heatcell{metricblue}{51}{49.67} & \heatcell{metricblue}{53}{63.63} & \heatcell{metricpurple}{56}{25.66} & \heatcell{metricpurple}{59}{20.70} & & \heatcell{metricteal}{47}{39.04} & \heatcell{metricteal}{50}{60.99} & \heatcell{metricgold}{49}{48.10} & \heatcell{metricgold}{56}{75.82} \\
\methodchip{4D9CAE}MSGSA~\cite{Lin2024MultiStageGeometricSemantic}           & \heatcell{metricblue}{56}{54.94} & \heatcell{metricblue}{54}{64.95} & \heatcell{metricpurple}{55}{25.28} & \heatcell{metricpurple}{58}{20.41} & &  \heatcell{metricgray}{8}{--}   &  \heatcell{metricgray}{8}{--}   &  \heatcell{metricgray}{8}{--}   &  \heatcell{metricgray}{8}{--}   \\
\methodchip{9372B2}BCLNet~\cite{Miao2024BCLNet}                              & \heatcell{metricblue}{53}{52.62} & \heatcell{metricblue}{55}{66.08} & \heatcell{metricpurple}{53}{24.59} & \heatcell{metricpurple}{57}{19.96} & & \heatcell{metricteal}{49}{40.56} & \heatcell{metricteal}{52}{62.71} & \heatcell{metricgold}{49}{48.07} & \heatcell{metricgold}{56}{75.84} \\
\baselineband \methodchip{476FB0}MatchMamba~\cite{Wu2025MatchMamba}                        & \heatcell{metricblue}{60}{\second{60.09}} & \heatcell{metricblue}{59}{\second{70.23}} & \heatcell{metricpurple}{59}{\second{27.23}} & \heatcell{metricpurple}{60}{\second{21.37}} & & \heatcell{metricteal}{54}{\second{44.12}} & \heatcell{metricteal}{55}{\second{64.43}} & \heatcell{metricgold}{54}{\second{52.19}} & \heatcell{metricgold}{58}{\second{77.71}} \\
\methodchip{476FB0}MatchMamba$^\dagger$~\cite{Wu2025MatchMamba}              & \heatcell{metricgray}{16}{60.09} & \heatcell{metricgray}{16}{67.60} & \heatcell{metricgray}{16}{25.70} & \heatcell{metricgray}{16}{20.97} & &  \heatcell{metricgray}{8}{--}   &  \heatcell{metricgray}{8}{--}   &  \heatcell{metricgray}{8}{--}   &  \heatcell{metricgray}{8}{--}   \\
\midrule
\ours \methodchip{C6384A}\oursname{SFMambaNet (ours)}                                   & \heatcell{metricblue}{62}{\best{62.28}} & \heatcell{metricblue}{62}{\best{73.83}} & \heatcell{metricpurple}{62}{\best{28.74}} & \heatcell{metricpurple}{62}{\best{21.97}} & & \heatcell{metricteal}{62}{\best{50.20}} & \heatcell{metricteal}{62}{\best{70.10}} & \heatcell{metricgold}{62}{\best{58.82}} & \heatcell{metricgold}{62}{\best{81.04}} \\
\bottomrule
\end{tabular}
\endgroup
\end{table*}

\subsection{Relative pose estimation}
\label{sec:pose}

\Cref{tab:pose} summarizes relative pose estimation.
SFMambaNet leads on every split, outdoors and indoors, with SIFT and with SuperPoint.
On unknown YFCC100M it reaches 73.83\% \mapfive, 3.60 points above the published MatchMamba and 6.23 points above our retrained one, and with SuperPoint the lead exceeds six points on both splits.
Indoors, where weak texture and repeated furniture leave few reliable local cues, it stays ahead on both SUN3D splits.
We find that, with either descriptor, the margin is larger on unknown outdoor scenes than on known ones: it grows where pruning has to generalize.
We read this as a direct benefit of the frequency view, since the coherence of correct motion is a property of geometry rather than of a particular scene.

\begin{figure}[t]
\centering
\includegraphics[width=\linewidth]{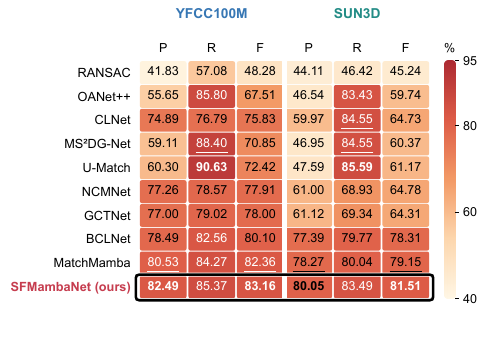}
\caption{\textbf{Outlier removal} with SIFT: precision (P), recall (R), and F-score (F) in \%. Best results are bold and second best are underlined; the complete table is in Appendix~E.}
\label{fig:outlier_profile}
\end{figure}

\begin{figure}[t]
\centering
\includegraphics[width=\linewidth]{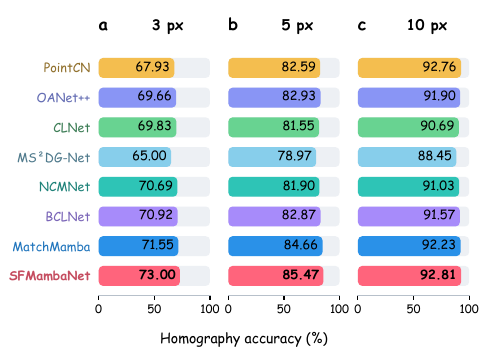}
\caption{\textbf{Homography estimation} on HPatches: accuracy (\%) at corner errors below 3, 5, and 10 pixels, for models trained on YFCC100M with SIFT.}
\label{fig:homography_bars}
\end{figure}

\subsection{Outlier removal}
\label{sec:outlier}

\begin{figure*}[t]
\centering
\setlength{\tabcolsep}{0pt}
\begin{tabular}{@{}p{0.248\textwidth}p{0.248\textwidth}p{0.248\textwidth}p{0.248\textwidth}@{}}
\centering\footnotesize Putative correspondences & \centering\footnotesize BCLNet~\cite{Miao2024BCLNet} & \centering\footnotesize MatchMamba~\cite{Wu2025MatchMamba} & \centering\arraybackslash\footnotesize SFMambaNet (ours)
\end{tabular}\\[1pt]
\includegraphics[width=\textwidth, trim=0 1669.24 0 320, clip]{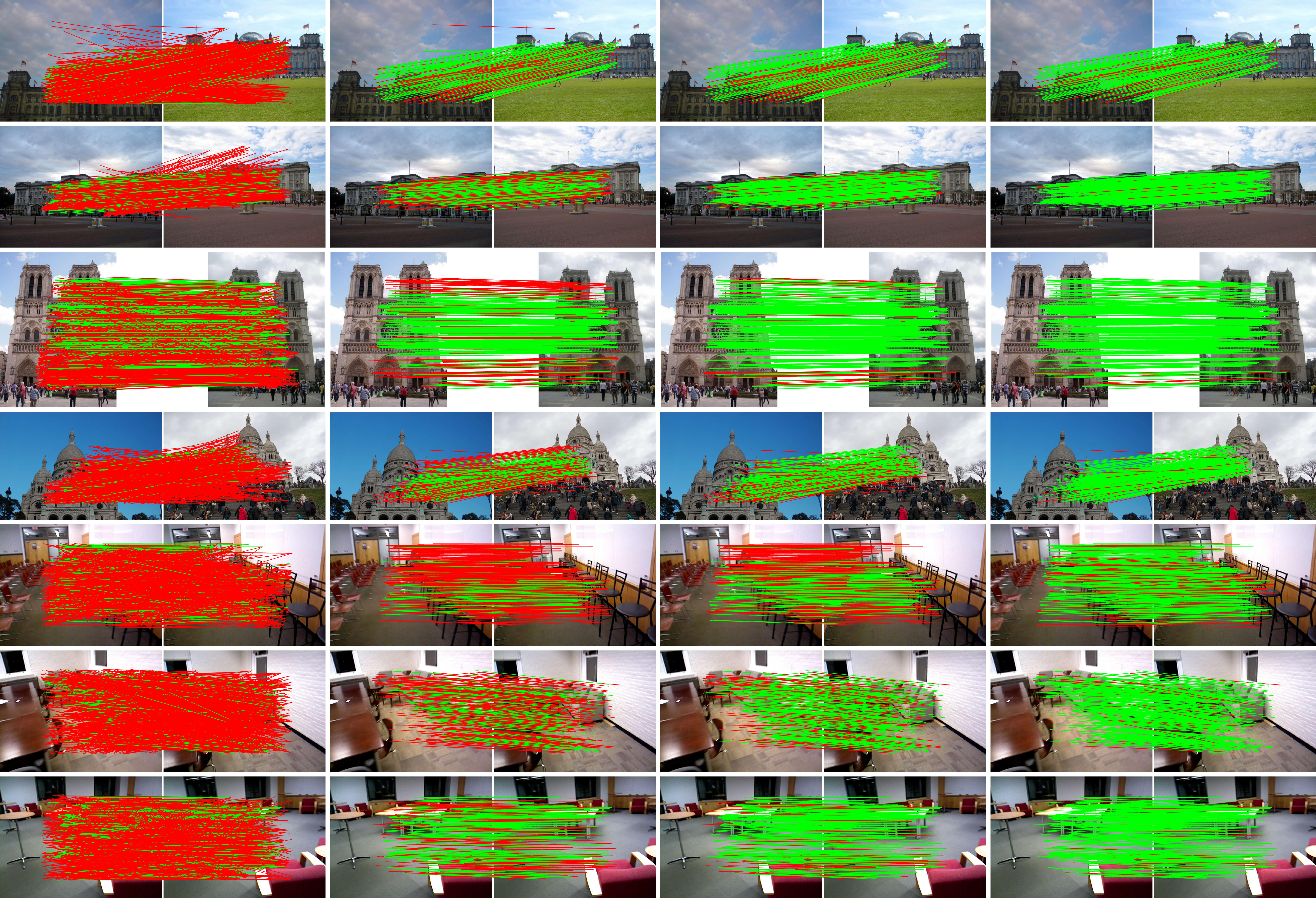}\\[-0.5pt]
\includegraphics[width=\textwidth, trim=0 969.24 0 1054, clip]{fig/fig_Outlier_Removal.pdf}\\[-0.5pt]
\includegraphics[width=\textwidth, trim=0 647.24 0 1344, clip]{fig/fig_Outlier_Removal.pdf}\\[-0.5pt]
\includegraphics[width=\textwidth, trim=0 323.24 0 1666, clip]{fig/fig_Outlier_Removal.pdf}
\caption{\textbf{Qualitative results} on the unknown splits of YFCC100M (top two rows) and SUN3D (bottom two rows). \inl{Green} and \outl{red} lines are inliers and outliers among the retained correspondences. More examples are given in Appendix~G.}
\label{fig:qual}
\end{figure*}

SFMambaNet is the most precise pruner on both datasets and also reaches the best F-score (\cref{fig:outlier_profile}), with its largest gain over MatchMamba indoors.
We find that SFMambaNet does not buy precision with recall: against MatchMamba it raises both, on both datasets, so that verification starts from a set that is cleaner and fuller at once.
\Cref{fig:qual} makes the difference visible on wide baselines, repeated structures, and weak texture: the competitors keep many outliers (\outl{red}), whereas SFMambaNet keeps dense, clean inliers (\inl{green}).

\subsection{Homography estimation}
\label{sec:homography}

To test generalization to planar scenes, we apply the models trained on YFCC100M, without fine-tuning, to the 116 HPatches sequences~\cite{Balntas2017HPatches} (up to 4000 SIFT keypoints per image) and report the fraction of pairs whose mean corner error is below 3, 5, and 10 pixels~\cite{DeTone2018SuperPoint}.
Although trained only to estimate essential matrices, SFMambaNet is the most accurate at every threshold (\cref{fig:homography_bars}; Appendix~F).
We observe that its lead over MatchMamba grows as the threshold tightens and peaks at 3 pixels: SFMambaNet does not merely find a usable homography more often, it finds a more precise one.

\subsection{Ablation study}
\label{sec:ablation}

We report means over five paired seeds ($\pm$ and error bars: sample SD), with SIFT, no RANSAC, identical training budgets, and validation-selected settings (Appendices~L--N).

\begin{table}[t]
\centering
\caption{Component additions on YFCC100M (mAP, \%; five-seed mean $\pm$ SD). The baseline replaces LSGA and SIGM by identities; SIGM uses the order that LSGA supplies.}
\label{tab:abl_main}
\setlength{\tabcolsep}{1.8pt}
\footnotesize
\renewcommand{\arraystretch}{1.12}
\medmuskip=2mu
\begin{tabular}{@{}ccccc@{}}
\toprule
\multirow{2}{*}[-\dimexpr(\aboverulesep+\belowrulesep+\cmidrulewidth)/2\relax]{Variant} & \multicolumn{2}{c}{\metricheader{Known}} & \multicolumn{2}{c}{\auxheader{Unknown}} \\
\cmidrule(lr){2-3}\cmidrule(lr){4-5}
 & \metricheader{$5^\circ$} & \metricheader{$20^\circ$} & \auxheader{$5^\circ$} & \auxheader{$20^\circ$} \\
\midrule
\tabbaseline Baseline & $40.42\tabunc{.57}$ & $61.18\tabunc{.36}$ & $55.87\tabunc{.48}$ & $75.13\tabunc{.31}$ \\
+ LSGA & $54.63\tabunc{.33}$ & $74.88\tabunc{.45}$ & $64.71\tabunc{.39}$ & $83.32\tabunc{.26}$ \\
\tabfinal \oursname{+ LSGA + SIGM} & $\tabbest{62.17}\tabunc{.43}$ & $\tabbest{79.36}\tabunc{.28}$ & $\tabbest{73.64}\tabunc{.37}$ & $\tabbest{87.08}\tabunc{.19}$ \\
\bottomrule
\end{tabular}
\end{table}

\paragraph{Main components.}
Starting from a baseline that keeps the pruning backbone, prediction heads, and verification, LSGA adds 8.84 points of unknown-split \mapfive and SIGM another 8.93 (\cref{tab:abl_main}).
We find that the two modules age differently: from known to unseen scenes, the gain of LSGA shrinks, whereas that of SIGM holds and at $5^\circ$ even grows, so the global frequency gating is what carries the model to new scenes.
Within LSGA, each added sub-component raises accuracy in turn (Appendix~L).
Encoding and gating, the two frequency components, are isolated below.

\paragraph{Gate and encoding controls.}
\Cref{fig:gate_control} swaps the learned gates for identities, for fixed Gaussian low-pass filters with validation-selected cutoffs,\footnote{Each gate position takes the best of five cutoffs on validation data, which gives the fixed filter its best chance (Appendix~L).} and for point-wise MLPs of equal size; the learned gates win on both datasets.
\Cref{tab:encoding_control} swaps the geometric encoding inside an unchanged attention, from raw coordinates and a linear lift to our Fourier features, which are clearly best; \cref{sec:analysis} explains why.

\begin{table*}[t]
\centering
\begin{minipage}[t]{\columnwidth}
\centering
\vspace{0pt}
\includegraphics[width=\linewidth]{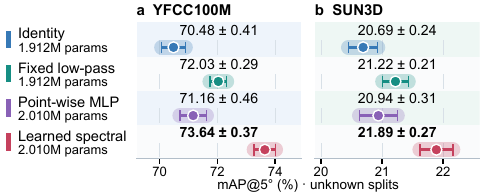}
\captionof{figure}{Gate controls on unknown splits (\mapfive, \%; five-seed mean $\pm$ SD; labels give parameter counts). All variants keep LSGA, both scans, residual fusion, and normalization.}
\label{fig:gate_control}
\label{tab:gate_control}
\label{tab:controlled_gates}
\end{minipage}\hfill
\begin{minipage}[t]{\columnwidth}
\centering
\captionof{table}{Encoding control under identical LGA and learned gates (five-seed \mapfive, \%). $d_{\rm in}$: encoding input dimension; every embedding outputs $D=128$ channels.}
\label{tab:encoding_control}
\label{tab:controlled_encoding}
\small
\setlength{\tabcolsep}{2.2pt}
\renewcommand{\arraystretch}{1.12}
\renewcommand{\tabularxcolumn}[1]{m{#1}}
\begin{tabularx}{\linewidth}{@{}>{\centering\arraybackslash}X >{\centering\arraybackslash}m{0.07\linewidth} >{\centering\arraybackslash}m{0.125\linewidth} >{\centering\arraybackslash}m{0.215\linewidth} >{\centering\arraybackslash}m{0.215\linewidth}@{}}
\toprule
\neutralheader{Encoding} & \neutralheader{$d_{\rm in}$} & \neutralheader{\makecell[c]{Params\\(M)}} & \metricheader{\makecell[c]{YFCC\\Unk.}} & \auxheader{\makecell[c]{SUN3D\\Unk.}} \\
\midrule
\tabbaseline None explicit & -- & 1.944 & $70.83\tabunc{.43}$ & $20.76\tabunc{.30}$ \\
Raw-coordinate MLP & 4 & 2.010 & $71.63\tabunc{.34}$ & $21.17\tabunc{.23}$ \\
Linear lift + MLP & 128 & 2.010 & $71.72\tabunc{.39}$ & $21.11\tabunc{.28}$ \\
\tabfinal \oursname{Fourier + MLP} & 128 & 2.010 & $\tabbest{73.64}\tabunc{.37}$ & $\tabbest{21.89}\tabunc{.27}$ \\
\bottomrule
\end{tabularx}
\end{minipage}
\end{table*}

\paragraph{Efficiency.}
With slightly fewer parameters and FLOPs and a shorter inference time, SFMambaNet improves on MatchMamba by 3.60 points of \mapfive (\cref{fig:efficiency_costs}; Appendix~J).
We find that frequency modeling is nearly free: the twelve spectral gates hold under 5\% of the parameters, and the whole model beats BCLNet~\cite{Miao2024BCLNet} with under a third of its parameters at more than twice its speed.

\section{Analysis}
\label{sec:analysis}

The benchmarks show that frequency modeling helps; we now ask why and test each answer by experiment.

\paragraph{Frequency, not capacity, drives the gain.}
If the gates helped merely by adding parameters, any layer of the same size should do as well.
We test this by replacing them with point-wise MLPs\lgMLP{} of exactly their parameter count, and most of the improvement disappears; parameter-free low-pass filters\lgLowpass{} keep more of it, and learning the response\lgLearned{} adds the rest (\cref{fig:gate_control}).
What matters is filtering along the cluster order.
Placement agrees: the forward gates alone, half of the total, recover more than two thirds of the gain (Appendix~L), as they shape what the second scan receives.

\paragraph{The gates learn to be low-pass.}
Nothing in the training objective asks for a particular frequency response, yet every trained gate attenuates all but its lowest few frequencies (Appendix~N).
The network relies on this shape rather than on energy: restoring only the high-frequency gains of the first gate costs more than four fifths of what removing the gate costs, whereas rescaling its output to the same energy costs nothing (Appendix~I).
Left to itself, the network thus passes on the slowly varying part of the cluster-ordered features, which is where, by our premise of smooth inliers and erratic outliers, agreement between matches lives.

\paragraph{Gating also tames the scan order.}
Mamba reads a sequence, but correspondences form a set, so a Mamba pruner should be sensitive to their order.
Along the cluster order, however, a shuffle only reorders correspondences inside each cluster, adding a zero-mean residual with little low-frequency energy (\cref{sec:sigm}), exactly what the gates attenuate.
The gated scan should thus barely notice a shuffle, and it does: shuffling all correspondences changes at most 1.5\% of those retained and the pose accuracy by less than a tenth of a point, several times less than with identity gates (Appendix~M).
A component added for accuracy thus also turns an order-dependent scan into a nearly order-agnostic one.

\paragraph{Sinusoids, not dimensions, expose local geometry.}
If Fourier features helped only by raising the input dimension, a linear lift to the same 128 dimensions should do as well.
It does not: the lift gains almost nothing over raw coordinates, whereas Fourier features roughly triple the gain that raw coordinates bring over no explicit geometry (\cref{tab:encoding_control}).
We attribute this to the spectral bias of MLPs~\cite{Rahaman2019SpectralBias,Tancik2020FourierFeatures}: the small offsets that separate an inlier from a nearby outlier yield nearly identical features until sinusoids of many frequencies pull them apart.
Fourier features and learned gates each help without the other (Appendix~L), so frequency modeling pays off at both the local and the global scale.

\section{Related Work}
\label{sec:related}

\paragraph{Correspondence pruning.}
Classical pipelines fit a model by sampling, as in RANSAC~\cite{Fischler1981RANSAC} and its successors~\cite{Chum2005PROSAC,Torr2000MLESAC,Raguram2012USAC,Barath2019MAGSAC,Barath2020MAGSACpp}, or enforce the coherence of correct motion~\cite{Ma2014VectorFieldConsensus,Bian2017GMS,Ma2019LocalityPreservingMatching,Fan2023SmoothnessDrivenConsensus,Xia2025GGSLC}.
LFGC~\cite{Yi2018GoodCorrespondences} turned pruning into per-correspondence classification with a permutation-equivariant network~\cite{Qi2017PointNet,Zaheer2017DeepSets}, which context normalization and clustering made stronger~\cite{Zhang2019OrderAwareNetwork,Sun2020ACNe,Zhong2021TNet}.
Graph networks then mined local consensus~\cite{Zhao2019NMNet,Zhao2021ProgressiveCorrespondencePruning,Dai2022MS2DGNet,Li2023UMatch,Liu2023PGFNet,Liu2023ProgressiveNeighborConsistencyMining,Dai2024MGNet,Lin2024MultiStageGeometricSemantic,Guo2024GraphContextTransformation,Li2025SGNNet,WangVol61PGNet,ZhuangVol219MultiGranularityConsensus}, attention modeled longer-range dependencies~\cite{Liao2023SGANet,Yang2024CGRNet,LiaoVol38VSFormer,LiuVol159TransMatch,Shen2024CLGNet,MGCA-Net,Miao2024BCLNet,Dai2026LeCoT}, and geometry-consistent pre-training improved generalization~\cite{Liao2024GCPretrain}.
Such pruners complement learned sparse and dense matchers~\cite{Sarlin2020SuperGlue,Sun2021LoFTR,Lindenberger2023LightGlue,Ryoo2025MambaGlue}.

\paragraph{State space models on unordered data.}
SSMs~\cite{Gu2021EfficientlyModelingLongSequences} and Mamba~\cite{Gu2023Mamba,Dao2024Mamba2} provide linear-time sequence context and now reach audio~\cite{Goel2022ItsRaw}, images and videos~\cite{Nguyen2022S4ND,Zhu2024VisionMamba,Liu2024VMamba,Huang2024LocalMamba}, medical images~\cite{Wang2024MambaUNet,Xing2024SegMamba}, point clouds~\cite{Liang2024PointMamba,Han2024Mamba3D}, and graphs~\cite{Behrouz2024GraphMambaKDD,Wang2024GraphMambaSelectiveSSM}.
Unordered inputs must first be serialized: point-cloud models follow space-filling curves~\cite{Liang2024PointMamba,Wu2024PTv3}, graph models rank nodes or sample walks~\cite{Wang2024GraphMambaSelectiveSSM,Behrouz2024GraphMambaKDD}, while attention is order-free but quadratic when global~\cite{Vaswani2017AttentionIsAllYouNeed,Dosovitskiy2020ViT}.
For pruning, MatchMamba~\cite{Wu2025MatchMamba} combines cluster-spatial and flip-bidirectional scans, MambaMatch~\cite{miao2025mambamatch} fuses full- and cluster-scale scans, and CorrMamba~\cite{Fang2026CorrMamba} learns a causal ordering.
Our order comes from learned clusters and has a frequency meaning.

\paragraph{Frequency-domain learning.}
Fourier features lift low-dimensional coordinates~\cite{Rahimi2007RFF,Tancik2020FourierFeatures,Mildenhall2020NeRF} against the spectral bias of neural networks~\cite{Rahaman2019SpectralBias}, global filters and Fourier mixing replace attention at $\mathcal{O}(N\log N)$ cost~\cite{Rao2021GFNet,Guibas2022AFNO,LeeThorp2022FNet}, graph signal processing defines frequencies on irregular domains~\cite{Shuman2013GSP}, and frequency cues accompany Mamba in image fusion~\cite{Sun2025SFMFusion} and camouflage detection~\cite{deng2025fmnetfrequencyassistedmambalikelinear}.
In pruning, smoothness priors~\cite{Liu2021LMCNet,Zhang2023ConvMatch,Zhang2024DeMatch,Xia2025GGSLC} regularize correspondence motion; SFMambaNet instead filters learned features between scans, along an order that its own clustering provides.

\section{Conclusion}
\label{sec:conclusion}
We study whether the contrast between smooth inliers and erratic outliers, which people read at a glance and robust matching has long relied on, can be built into the scan of a Mamba pruner.
By scanning correspondences cluster by cluster and gating frequencies between scans,\footnote{We discuss the limitations of our setup in Appendix~A.} we show that SFMambaNet outperforms the compared pruners on all three tasks with about two million parameters.
Specifically, it leads most on unseen scenes, raises precision and recall together, and yields the most precise homographies, although trained only to estimate essential matrices.
We further find that the gain comes from filtering rather than capacity, and that the learned gates turn low-pass on their own.
Our analysis of scan order supports one explanation: a shuffle only perturbs high-frequency residuals inside clusters, which the gates suppress.
Beyond the global scan, we find that Fourier features help the local attention through their frequencies, not their dimension, so frequency pays off at both scales.
Together, these results show that frequency is a natural language for describing agreement between correspondences, and that a Mamba scan over learned clusters can treat an unordered set of matches consistently and efficiently.

\section*{LLM Usage Statement}
In this work, we used generative AI tools for providing feedback on research methodology and implementing methods.
We have not used generative AI tools for proposing or refining hypotheses, generating synthetic datasets or cleaning and reformatting existing datasets, helping to develop theoretical models or conceptual frameworks, supporting qualitative and thematic data analysis, or interpreting the results.
Formulating mathematical claims, assisting in the writing of proofs, and assisting with translation are not applicable to this work.
Additionally, we used generative AI tools for creating or editing software code, drafting parts of a research paper, editing a research paper to improve readability, and identifying relevant literature.
We have reviewed all AI-assisted work.
In particular, LLM-generated code was verified and tested for correctness; all parts of the paper were ultimately written by the authors; and relevant literature identified by LLMs was used only for the authors to review and subsequently discuss in the paper.
We take responsibility for the final content of this work, including text, claims, or artifacts produced with the aid of generative AI.

{\small
  \raggedright
  \bibliographystyle{ieeenat_fullname}
  \bibliography{references}
}
\clearpage
\appendix
\setcounter{figure}{0}
\setcounter{table}{0}
\setcounter{equation}{0}
\setcounter{algorithm}{0}
\renewcommand{\thefigure}{S\arabic{figure}}
\renewcommand{\thetable}{S\arabic{table}}
\renewcommand{\theequation}{S\arabic{equation}}
\renewcommand{\thealgorithm}{S\arabic{algorithm}}
\renewcommand{\theHfigure}{appendix.\arabic{figure}}
\renewcommand{\theHtable}{appendix.\arabic{table}}
\renewcommand{\theHequation}{appendix.\arabic{equation}}
\renewcommand{\theHalgorithm}{appendix.\arabic{algorithm}}
\makeatletter
\renewcommand{\theHALG@line}{appendix.\arabic{ALG@line}}
\makeatother
\twocolumn[{\raggedright\Large\bfseries Appendix\par\vspace{12pt}}]
\section{Limitations}
\label{sec:s_limits}

Our setup leaves a few promising directions open; we briefly describe each below.

First, SFMambaNet treats agreement as low-frequency structure along the cluster order.
This premise also serves scenes that challenge it: indoors, where repeated furniture and weak texture leave few distinctive cues, SFMambaNet leads on both SUN3D splits (Tab.~\ref*{tab:pose} of the main paper), and its F-score gain over MatchMamba is about three times that outdoors (\cref{sec:s_tables}).
Strongly repeated patterns can still yield groups of mismatches that agree with one another, a case shared by all consistency-based pruners.
Reasoning explicitly about competing motions inside a cluster is a natural extension.

Second, \cref{sec:s_order} analyzes one pruning stage and one gate exactly, and \cref{sec:s_local_permutations} tests the complete two-stage network directly: under random permutations of its input, the learned gates reduce the mean change of \mapfive from 0.62 to 0.08 points on YFCC100M and from 0.42 to 0.06 points on SUN3D, relative to identity gates (\cref{tab:input_permutations}).
An architecture whose invariance is guaranteed end to end is a natural next step.

Third, following the protocol of prior pruning networks~\cite{Zhang2019OrderAwareNetwork,Zhao2021ProgressiveCorrespondencePruning,Wu2025MatchMamba}, we report the pose accuracy of the differentiable eight-point estimate without RANSAC.
Combining the pruned correspondences with a robust estimator such as MAGSAC~\cite{Barath2019MAGSAC} is a natural step for applications.

\begingroup
\suppalgorithmblue
\begin{algorithm*}[t]
\caption{SFMambaNet inference and its LSGA/SIGM operators}
\label{alg:s_inference}
\begin{suppalgorithmcontent}
\begin{algorithmic}[1]
\Require Normalized correspondences $C\in\mathbb{R}^{N\times4}$, trained parameters, verification threshold $\tau$
\Ensure Estimated essential matrix $\hat{\mathbf{E}}$ and verified correspondence set $C_{\mathrm{in}}$
\Function{Infer}{$C,\tau$}
  \State $(\mathcal{W}_1,C_1)\gets f_{\theta_1}(C)$ \Comment{First pruning block}
  \State $(\mathcal{W}_2,C_2)\gets f_{\theta_2}(\mathcal{W}_1,C_1)$ \Comment{Second pruning block}
  \State $\hat{\mathbf{E}}\gets g(\mathcal{W}_2,C_2)$ \Comment{Differentiable weighted eight-point algorithm}
  \State $\mathcal{D}\gets h(\hat{\mathbf{E}},C)$; $C_{\mathrm{in}}\gets\{c_i\in C:\mathcal{D}_i<\tau\}$ \Comment{Verify \textbf{all} original correspondences}
  \State \Return $\hat{\mathbf{E}},C_{\mathrm{in}}$
\EndFunction
\Statex
\Function{LSGA}{$C,F^F,\mathcal{N}$}
  \State $\Delta C_{:,i,j}\gets(C_{\mathcal{N}(i,j),:}-C_{i,:})^{\top}$ for all $i,j$; gather $F^F_{\mathcal{N}}$ from $F^F$
  \State $\alggeo{E_{\mathit{freq}}}\gets[\sin(2\pi\mathbf{B}\Delta C);\cos(2\pi\mathbf{B}\Delta C)]$ \Comment{\alggeo{\textbf{Fourier geometry}}; fixed $\mathbf{B}$}
  \State $\alggeo{E_{\mathit{geo}}}\gets\MLP(E_{\mathit{freq}})$
  \State $Q\gets W_QF^F$; $K\gets W_K(F^F_{\mathcal{N}}+E_{\mathit{geo}})$; $V\gets W_V(F^F_{\mathcal{N}}+E_{\mathit{geo}})$
  \State $\mathcal{A}\gets\Softmax_k\big(\BMM(Q^{\top},K)/\sqrt{D}\big)$; $F^A\gets\BMM(V,\mathcal{A}^{\top})$
  \State $F^M\gets\LN\big(\MLP(F^F+F^A)\big)$
  \State $(G^C,S)\gets\operatorname{DiffPool}(F^M)$; $\algord{\Pi}\gets\operatorname{Order}(S)$ \Comment{\algord{\textbf{Cluster-induced order}}}
  \State $F^G\gets\algord{\Pi^{-1}}\big(\operatorname{BiMamba}(\algord{\Pi}(F^M))\big)$
  \State $F^L\gets\operatorname{BiCSM}(G^C)$ \Comment{Cluster spatial Mamba}
  \State $F^I\gets F^L\Softmax_M\big((F^L)^{\top}F^G/\sqrt{D}\big)$; $F^{GL}\gets\LN(F^M+F^I)$
  \State \Return $F^{GL},\algord{\Pi}$
\EndFunction
\Statex
\Function{SIGM}{$X,\Pi$}
  \State $X_{\mathit{ord}}\gets\algord{\Pi}(X)$ \Comment{Reuse the LSGA order}
  \State $F_{\mathit{fwd}}\gets\operatorname{Mamba}(X_{\mathit{ord}})$
  \State $F^S_{\mathit{fwd}}\gets\Ifft\big(\Fft(F_{\mathit{fwd}})\odot\Interp(\algspec{W^1_{\mathit{gate}}})\big)$ \Comment{\algspec{\textbf{Forward spectral gate}}}
  \State $F_{\mathit{bwd}}\gets\operatorname{Mamba}\big(\Flip(F^S_{\mathit{fwd}})\big)$ \Comment{Scan the \textbf{gated} forward features}
  \State $F^S_{\mathit{bwd}}\gets\Ifft\big(\Fft(F_{\mathit{bwd}})\odot\Interp(\algspec{W^2_{\mathit{gate}}})\big)$ \Comment{\algspec{\textbf{Backward spectral gate}}}
  \State $Y_{\mathit{ord}}\gets\LN\big(X_{\mathit{ord}}+F^S_{\mathit{fwd}}+\Flip(F^S_{\mathit{bwd}})\big)$
  \State \Return $\algord{\Pi^{-1}}(Y_{\mathit{ord}})$ \Comment{\algord{\textbf{Restore correspondence order}}}
\EndFunction
\end{algorithmic}
\end{suppalgorithmcontent}
\end{algorithm*}
\endgroup

\section{Local geometric attention in detail}
\label{sec:s_lga}

\begin{figure}[t]
  \centering
  \includegraphics[width=\linewidth]{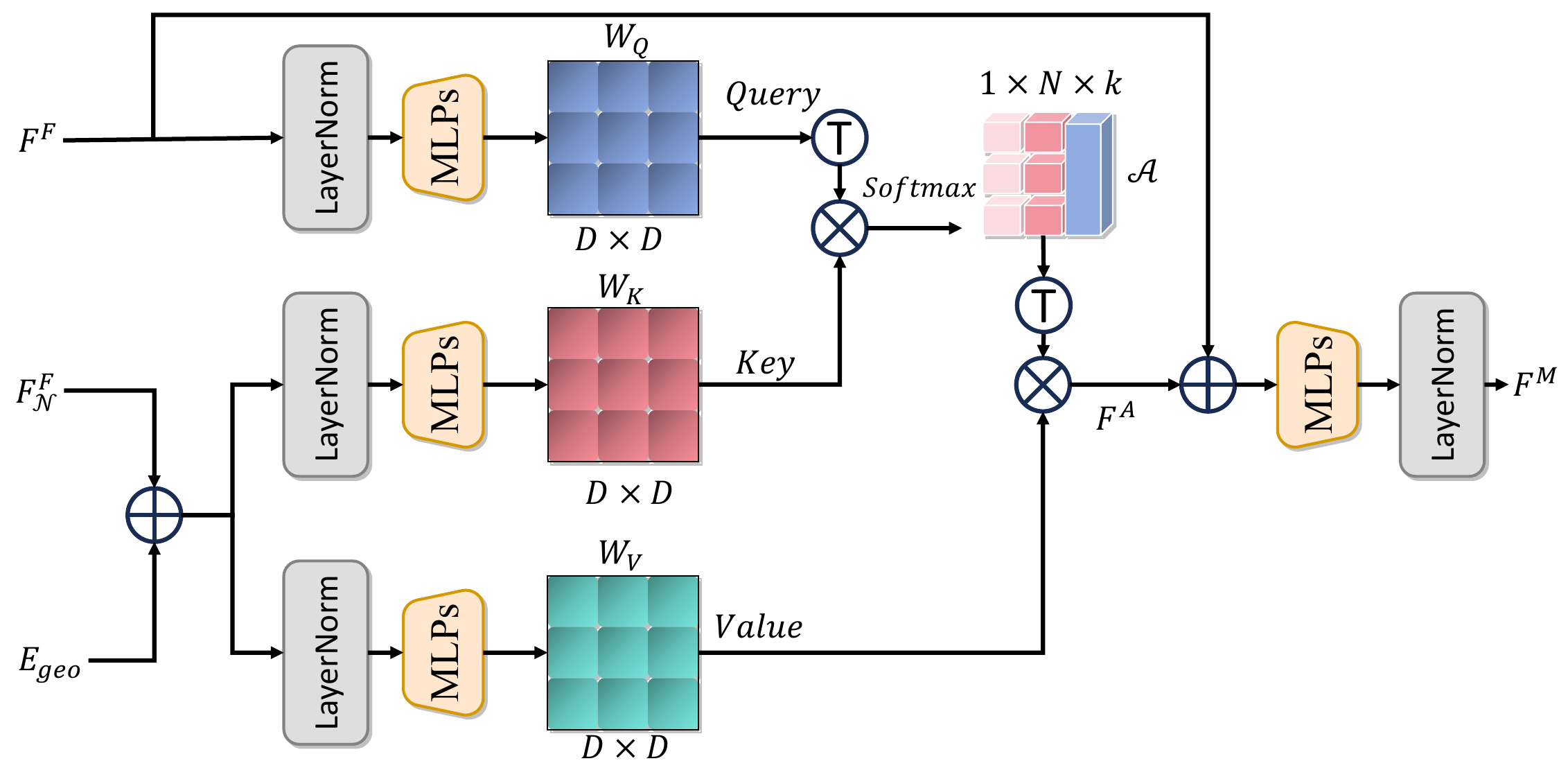}
  \caption{\textbf{Local geometric attention (LGA).} The central feature $F^F$ ($D\times N\times1$) forms the query, while the neighbor features $F^F_{\mathcal{N}}$ ($D\times N\times k$) are fused with the spectral-geometric embedding $E_{\mathit{geo}}$ to form keys and values. The attention map $\mathcal{A}$ ($1\times N\times k$) is obtained by projecting the query onto the keys along the channel dimension, and the aggregated feature $F^A$ weights the values with $\mathcal{A}$; a residual MLP and layer normalization produce $F^M$.}
  \label{fig:s_lga}
\end{figure}

\Cref{fig:s_lga} shows the local geometric attention of LSGA (Eqs.~(\ref*{eq:qkv}) and (\ref*{eq:lga}) of the main paper).
For two tensors $X\in\mathbb{R}^{a\times N\times b}$ and $Y\in\mathbb{R}^{b\times N\times c}$ we define the $N$-wise batched matrix product
\begin{equation}
  \big[\operatorname{BMM}_N(X,Y)\big]_{:,i,:}=X_{:,i,:}\,Y_{:,i,:},\qquad i=1,\dots,N,
\end{equation}
so that $\operatorname{BMM}_N(X,Y)\in\mathbb{R}^{a\times N\times c}$.
The transpose $Q^{\top}\in\mathbb{R}^{1\times N\times D}$ swaps the first and third dimensions of $Q\in\mathbb{R}^{D\times N\times1}$; consequently $\operatorname{BMM}_N(Q^{\top},K)\in\mathbb{R}^{1\times N\times k}$ contains, for every correspondence, the dot products between its query and the keys of its $k$ neighbors.
The softmax is taken along the neighbor dimension, so that each correspondence distributes its attention over its own neighborhood only, and $F^A=\operatorname{BMM}_N(V,\mathcal{A}^{\top})\in\mathbb{R}^{D\times N\times1}$ with $\mathcal{A}^{\top}\in\mathbb{R}^{k\times N\times1}$.
The cost is $\mathcal{O}(NkD)$, compared with $\mathcal{O}(N^2D)$ for full pairwise attention.
The multiplication of the projection matrix $\mathbf{B}\in\mathbb{R}^{d_f\times4}$ with the relative-coordinate tensor $\Delta C\in\mathbb{R}^{4\times N\times k}$ in Eq.~(\ref*{eq:sge}) is performed along the first dimension, \ie, the projection is shared by all $N\times k$ neighbor pairs.

\section{Implementation details}
\label{sec:s_impl}

\subsection{Architecture and training}

SFMambaNet is implemented in PyTorch and processes $N=2000$ putative correspondences obtained from SIFT or SuperPoint descriptors.
Each of the two pruning blocks retains half of its input (Fig.~\ref*{fig:arch}).
Its feature extractor consists of an up-projection, four ResNet blocks, the local graph block, four ResNet blocks, and one LSGA block; its context aggregator consists of four ResNet blocks, three SIGM blocks, and four ResNet blocks.
The channel dimension is $D=128$, the Fourier projection dimension is $d_f=64$, the number of clusters is $M=256$, and the number of neighbors is $k=18$.
The state size of the Mamba layers is 8 in the LSGA block and 16 in the SIGM blocks.
The complex gates $W^1_{\mathit{gate}}$ and $W^2_{\mathit{gate}}$ are resampled by $\operatorname{Interp}(\cdot)$ to the number of frequencies of the current sequence, because the two pruning blocks operate on sequences of different lengths.

We optimize with Adam~\cite{Kingma2014Adam}, a batch size of 32, and 500k iterations.
The learning rate is warmed up linearly from an initial value of $10^{-3}$ during the first 10k iterations and is then decayed by a factor of 0.4 every 20k iterations.
Following CLNet~\cite{Zhao2021ProgressiveCorrespondencePruning}, correspondences with a symmetric epipolar distance below $10^{-4}$ are labeled as inliers.
The weight $\gamma$ of the geometric loss in Eq.~(\ref*{eq:loss}) is 0 for the first 20k iterations, so that the classification is learned first, and 0.5 afterwards.
All experiments were run on a single NVIDIA RTX~3090 GPU under Ubuntu~22.04.

\subsection{Datasets and loss}
\paragraph{Datasets.}
Following OANet~\cite{Zhang2019OrderAwareNetwork}, YFCC100M~\cite{Thomee2016YFCC100M} uses 68 training and 4 test sequences, and SUN3D~\cite{Xiao2013SUN3D} 239 training and 15 test sequences.
Training sequences are further divided into 60\%/20\%/20\% training, validation, and test pairs; their test pairs form the known split, whereas the held-out sequences form the unknown split.
Pose accuracy is the mAP of the angular rotation and translation errors at $5^\circ$ and $20^\circ$~\cite{Zhang2019OrderAwareNetwork}.

\paragraph{Geometric loss.}
$\mathcal{L}_{\mathit{geo}}$ in Eq.~(\ref*{eq:loss}) of the main paper regresses $\hat{\mathbf{E}}$ towards the ground truth through the Sampson distance of virtual correspondences $(\mathbf{p}_i,\mathbf{p}'_i)$ generated from $\mathbf{E}_{gt}$:
\begin{equation}
  \mathcal{L}_{\mathit{geo}}=\frac{\big(\mathbf{p}'^{\top}_i\hat{\mathbf{E}}\mathbf{p}_i\big)^2}{\sum_{j=1}^{2}\Big(\big\|\mathbf{E}_{gt}\mathbf{p}_i\big\|_{[j]}^2+\big\|\mathbf{E}_{gt}^{\top}\mathbf{p}'_i\big\|_{[j]}^2\Big)},
  \label{eq:s_geo}
\end{equation}
where $\|\cdot\|_{[j]}$ denotes the $j$-th element of a vector.

\subsection{Inference pseudocode}
Algorithm~\ref{alg:s_inference} summarizes SFMambaNet inference and its LSGA/SIGM operations.
The pruning blocks $f_{\theta_1},f_{\theta_2}$ include the feature extractor, context aggregator, and weight prediction; each retains half its input.
The LSGA routine takes the graph features $F^F$ and neighbor indices $\mathcal{N}$, and returns both the refined features and their cluster order $\Pi$.
The SIGM routine describes one module applied to its current input $X$; the context aggregator contains three such modules with their own parameters, bracketed by the ResNet groups described above.
All omitted projections and Mamba parameters belong to the corresponding module.
Blue highlights geometric encoding, indigo marks ordering and restoration, and rose marks spectral gates, as in the figures of the main paper; bold labels also identify these operations.

Here, $\operatorname{Order}(S)$ groups correspondences by their strongest cluster slot.
The softmax in the multi-scale interaction is taken over the $M$ clusters for each correspondence.
The FFT and $\Flip$ operate along the sequence axis, and $\Interp$ adapts each gate to its current length with the real-output constraints stated in \cref{sec:s_controlled_definitions}.
The second $\Flip$ restores the backward direction before residual fusion; $\Pi^{-1}$ then restores the correspondence order.

\section{Properties of the cluster-induced order}
\label{sec:s_order}

Sec.~\ref*{sec:sigm} of the main paper states two properties of the order $\Pi$ along which SIGM applies its Fourier transform.
We make them precise here.
Throughout, $\Pi$ arranges the correspondences in contiguous blocks $B_1,\dots,B_M$, one per cluster slot, where $B_m$ contains the correspondences whose strongest assignment in $S$ is slot $m$, and $L_{\max}=\max_m|B_m|$.
We use the unitary discrete Fourier transform $\hat r(\omega)=N^{-1/2}\sum_n r_n e^{-2\pi\mathrm{i}\,\omega n/N}$ with integer frequencies $-N/2<\omega\le N/2$, so that $\sum_\omega|\hat r(\omega)|^2=\|r\|_2^2$.

\begin{proposition}[Stability under input permutations]
\label{prop:stability}
Let $P$ permute the same input correspondence and feature set at a fixed pruning stage.
If the features $F^M$ and the network that maps them to the assignment $S$ are permutation-equivariant, then $S(PC)=P\,S(C)$.
Hence every block $B_m$ contains the same correspondences for $C$ and $PC$, the blocks appear in the same order, and the two sequences differ only by permutations inside the blocks.
\end{proposition}

\begin{proof}
All layers before the pooling, \ie, shared point-wise layers with set-level normalization, the $k$-nearest-neighbor graph, and the local attention, are permutation-equivariant. Viewing $F^M$ as a $D\times N$ matrix, this gives $F^M(PC)=F^M(C)P^{\top}$.
The same holds for the assignment network, so $S(PC)=P\,S(C)$, and the strongest slot of every correspondence is unchanged.
Grouping by slot index therefore yields the same blocks in the same order, and only the positions inside a block may change.
\end{proof}

Proposition~\ref{prop:stability} holds whenever the stage receives the same feature set up to a row permutation, with ties in distances and assignments broken consistently.
For the complete two-stage network, \cref{sec:s_local_permutations} measures the effect of permuting the network input directly (\cref{tab:input_permutations}).

\begin{proposition}[Scale separation]
\label{prop:scale}
Let $\{0,\dots,N-1\}$ be partitioned into contiguous blocks of at most $L$ positions, and let $r\in\mathbb{R}^N$ have zero sum on every block.
Then, for every cutoff $\omega_c\ge0$,
\begin{equation}
  \sum_{|\omega|\le\omega_c}|\hat r(\omega)|^2\;\le\;\Big(\frac{\pi\,\omega_c\,L}{N}\Big)^{2}\,\|r\|_2^2 ,
  \label{eq:s_bound}
\end{equation}
\ie, at most a fraction $(\pi\omega_cL/N)^2$ of the energy of $r$ lies at frequencies up to $\omega_c$.
\end{proposition}

\begin{proof}
Let $W$ be the space of real signals whose transform vanishes for $|\omega|>\omega_c$ and $P_W$ the orthogonal projection onto it, so that the left-hand side of \cref{eq:s_bound} equals $\|P_Wr\|_2^2$.
For $u\in\mathbb{R}^N$, let $\bar u$ denote the signal in which every entry of $u$ is replaced by the mean of $u$ over its block.
Since $r$ has zero sum on every block, $\langle\bar u,r\rangle=0$, and therefore
\begin{equation}
\begin{aligned}
  \|P_Wr\|_2&=\max_{u\in W,\,\|u\|_2=1}\langle u,r\rangle=\max_{u\in W,\,\|u\|_2=1}\langle u-\bar u,r\rangle\\
  &\le\|r\|_2\max_{u\in W,\,\|u\|_2=1}\|u-\bar u\|_2 .
\end{aligned}
\label{eq:s_dual}
\end{equation}
Two classical inequalities bound the last factor.
First, the Laplacian of a path with $\ell\le L$ nodes has the smallest non-zero eigenvalue $4\sin^2(\pi/2\ell)\ge4/L^2$, so the Poincar\'e inequality on every block $B$ reads $\sum_{n\in B}(u_n-\bar u_n)^2\le\frac{L^2}{4}\sum_{n,n+1\in B}(u_{n+1}-u_n)^2$.
Summing over the blocks, and adding the non-negative differences across block boundaries on the right, gives $\|u-\bar u\|_2^2\le\frac{L^2}{4}\sum_n(u_{n+1}-u_n)^2$ with cyclic differences.
Second, the difference operator multiplies $\hat u(\omega)$ by $e^{2\pi\mathrm{i}\,\omega/N}-1$, whose modulus $2|\sin(\pi\omega/N)|$ is at most $2\pi\omega_c/N$ on the support of $\hat u$, so $\sum_n(u_{n+1}-u_n)^2\le(2\pi\omega_c/N)^2\|u\|_2^2$ for $u\in W$.
Together, $\|u-\bar u\|_2\le(\pi\omega_cL/N)\,\|u\|_2$, which with \cref{eq:s_dual} proves the claim.
\end{proof}

Proposition~\ref{prop:scale} applies to the deviations of the cluster-ordered features from their block means, which have zero sum on every block by construction: the band well below $N/(\pi L_{\max})$ contains almost none of their energy.
In the trained models, the largest blocks hold at most 24 to 48 correspondences, depending on dataset and stage (\cref{fig:s_cluster_occupancy}); even for these conservative block lengths, the bound at $\omega_c=5$ is 9.9--22.2\%, whereas the measured residuals keep at most 0.60--1.40\% of their energy at these frequencies, an order of magnitude less (\cref{tab:s_residual_bounds}).
The same residual bound controls how a frequency-domain gate responds to a reordering inside the clusters.

\begin{corollary}[Insensitivity of the gate to the order inside clusters]
\label{cor:gate}
Let $x'$ be obtained from a channel $x$ of the cluster-ordered features by permuting entries inside the blocks, and let $y=\mathcal{F}^{-1}(G\odot\mathcal{F}x)$ and $y'=\mathcal{F}^{-1}(G\odot\mathcal{F}x')$ for a gate $G$ with $|G(\omega)|\le G_{\max}$ for all $\omega$ and $|G(\omega)|\le\varepsilon$ for $|\omega|>\omega_c$.
Then
\begin{equation}
  \|y'-y\|_2\le\Big(\frac{\pi\,G_{\max}\,\omega_c\,L_{\max}}{N}+\varepsilon\Big)\,\|x'-x\|_2 .
  \label{eq:s_gate}
\end{equation}
\end{corollary}

\begin{proof}
The difference $d=x'-x$ has zero sum on every block, because permuting entries inside a block preserves its sum.
By Parseval's identity and Proposition~\ref{prop:scale},
\begin{equation}
\begin{aligned}
  \|y'-y\|_2^2&=\sum_\omega|G(\omega)|^2\,|\hat d(\omega)|^2\\
  &\le G_{\max}^2\!\!\sum_{|\omega|\le\omega_c}\!\!|\hat d(\omega)|^2+\varepsilon^2\!\!\sum_{|\omega|>\omega_c}\!\!|\hat d(\omega)|^2\\
  &\le\Big(G_{\max}^2\Big(\frac{\pi\omega_cL_{\max}}{N}\Big)^{2}+\varepsilon^2\Big)\|d\|_2^2 ,
\end{aligned}
\label{eq:s_gate_quad}
\end{equation}
and \cref{eq:s_gate} follows from $\sqrt{a^2+b^2}\le a+b$ for $a,b\ge0$.
\end{proof}

In words: a gate that keeps little gain beyond a cutoff well below $N/(\pi L_{\max})$ passes almost none of a reordering inside the clusters.
The bound compares the gate outputs $T_GQx$ and $T_Gx$ at the same sequence positions, with $T_G=\mathcal{F}^{-1}\operatorname{diag}(G)\mathcal{F}$ and $Q$ permuting positions inside blocks; \cref{sec:s_conditional_diagnostics} measures this ratio for the trained gates, and \cref{sec:s_local_permutations} measures the correspondence-wise change of complete modules and of the complete network.
The block means form a piecewise-constant signal that no permutation inside the blocks can change, whatever its spectrum, so only the within-block residual needs a bound.

\section{Complete comparison tables}
\label{sec:s_tables}

\Cref{tab:s_outlier} extends Fig.~\ref*{fig:outlier_profile} of the main paper with all baselines.
The complete relative-pose results are given in Tab.~\ref*{tab:pose} of the main paper.
SFMambaNet raises precision and recall over MatchMamba on both datasets; indoors, its F-score gain is about three times that outdoors (2.36 versus 0.80 points).

\paragraph{Relation to other Mamba pruners.}
All prior results in Tab.~\ref*{tab:pose} of the main paper and in \cref{tab:s_outlier} come from the unified evaluation of MatchMamba~\cite{Wu2025MatchMamba} under the protocol of \cref{sec:s_impl}.
SFMambaNet shares building blocks with two other Mamba pruners: its multi-scale interaction follows the cross-scale fusion of MambaMatch~\cite{miao2025mambamatch}, and its cluster branch uses the BiCSM of MatchMamba.
On top of a model that already contains both, the learned frequency gates add 3.16 points of \mapfive on unknown YFCC100M and 1.20 points on unknown SUN3D over identity gates (Fig.~\ref*{fig:gate_control} of the main paper), and the complete SIGM adds 8.93 points on unknown YFCC100M (Tab.~\ref*{tab:abl_main} of the main paper).
CorrMamba~\cite{Fang2026CorrMamba} learns a causal ordering of the correspondences; SFMambaNet instead derives its order from learned clusters, which gives the scan a frequency meaning (\cref{sec:s_order}) that the gates exploit.

\begin{table}[t]
\centering
\caption{\textbf{Outlier removal} with SIFT on YFCC100M and SUN3D: precision (P), recall (R), and F-score (F) in \%. Prior results are taken from~\cite{Wu2025MatchMamba}. Darker cells indicate higher values within each metric column.}
\label{tab:s_outlier}
\setlength{\tabcolsep}{2.6pt}
\small
\begingroup
\renewcommand{\arraystretch}{1.05}
\renewcommand{\best}[1]{\textcolor{bestc}{\textbf{#1}}}
\renewcommand{\second}[1]{\textcolor{secondc}{\underline{#1}}}
\begin{tabular}{@{}c ccc ccc@{}}
\toprule
\multicolumn{1}{@{}c}{\multirow{2}{*}[-\dimexpr(\aboverulesep+\belowrulesep+\cmidrulewidth)/2\relax]{Method}} & \multicolumn{3}{c}{\metricheader{YFCC100M}} & \multicolumn{3}{c}{\altmetricheader{SUN3D}} \\
\cmidrule(lr){2-4}\cmidrule(lr){5-7}
 & \metricheader{P} & \metricheader{R} & \metricheader{F} & \altmetricheader{P} & \altmetricheader{R} & \altmetricheader{F} \\
\midrule
\methodchip{8F9AA8}RANSAC~\cite{Fischler1981RANSAC}                          & \heatcell{metricblue}{12}{41.83} & \heatcell{metricblue}{12}{57.08} & \heatcell{metricblue}{12}{48.28} & \heatcell{metricpurple}{12}{44.11} & \heatcell{metricpurple}{12}{46.42} & \heatcell{metricpurple}{12}{45.24} \\
\methodchip{8D9C68}PointNet++~\cite{Qi2017PointNetPP}                        & \heatcell{metricblue}{20}{48.42} & \heatcell{metricblue}{18}{61.16} & \heatcell{metricblue}{20}{54.05} & \heatcell{metricpurple}{14}{45.64} & \heatcell{metricpurple}{59}{83.43} & \heatcell{metricpurple}{31}{59.00} \\
\methodchip{A48668}LFGC~\cite{Yi2018GoodCorrespondences}                     & \heatcell{metricblue}{26}{53.12} & \heatcell{metricblue}{54}{85.51} & \heatcell{metricblue}{37}{65.53} & \heatcell{metricpurple}{16}{47.24} & \heatcell{metricpurple}{59}{83.45} & \heatcell{metricpurple}{33}{60.32} \\
\methodchip{686FA7}OANet++~\cite{Zhang2019OrderAwareNetwork}                 & \heatcell{metricblue}{29}{55.65} & \heatcell{metricblue}{55}{85.80} & \heatcell{metricblue}{40}{67.51} & \heatcell{metricpurple}{15}{46.54} & \heatcell{metricpurple}{59}{83.43} & \heatcell{metricpurple}{32}{59.74} \\
\methodchip{5E9C98}ACNe~\cite{Sun2020ACNe}                                   & \heatcell{metricblue}{28}{54.56} & \heatcell{metricblue}{56}{86.92} & \heatcell{metricblue}{39}{67.04} & \heatcell{metricpurple}{15}{46.44} & \heatcell{metricpurple}{60}{84.23} & \heatcell{metricpurple}{32}{59.87} \\
\methodchip{A27292}T-Net~\cite{Zhong2021TNet}                                & \heatcell{metricblue}{31}{57.48} & \heatcell{metricblue}{59}{88.38} & \heatcell{metricblue}{43}{69.66} & \heatcell{metricpurple}{16}{46.94} & \heatcell{metricpurple}{61}{84.53} & \heatcell{metricpurple}{33}{60.36} \\
\methodchip{72A877}CLNet~\cite{Zhao2021ProgressiveCorrespondencePruning}     & \heatcell{metricblue}{53}{74.89} & \heatcell{metricblue}{41}{76.79} & \heatcell{metricblue}{51}{75.83} & \heatcell{metricpurple}{34}{59.97} & \heatcell{metricpurple}{61}{84.55} & \heatcell{metricpurple}{39}{64.73} \\
\methodchip{6198B7}MS$^2$DG-Net~\cite{Dai2022MS2DGNet}                       & \heatcell{metricblue}{33}{59.11} & \heatcell{metricblue}{59}{88.40} & \heatcell{metricblue}{44}{70.85} & \heatcell{metricpurple}{16}{46.95} & \heatcell{metricpurple}{61}{84.55} & \heatcell{metricpurple}{33}{60.37} \\
\methodchip{A57E61}PGFNet~\cite{Liu2023PGFNet}                               & \heatcell{metricblue}{31}{57.54} & \heatcell{metricblue}{59}{88.77} & \heatcell{metricblue}{43}{69.82} & \heatcell{metricpurple}{16}{47.05} & \heatcell{metricpurple}{61}{\second{85.02}} & \heatcell{metricpurple}{33}{60.58} \\
\methodchip{A589BC}U-Match~\cite{Li2023UMatch}                               & \heatcell{metricblue}{35}{60.30} & \heatcell{metricblue}{62}{\best{90.63}} & \heatcell{metricblue}{47}{72.42} & \heatcell{metricpurple}{17}{47.59} & \heatcell{metricpurple}{62}{\best{85.59}} & \heatcell{metricpurple}{34}{61.17} \\
\methodchip{2F8C8A}NCMNet~\cite{Liu2023ProgressiveNeighborConsistencyMining} & \heatcell{metricblue}{56}{77.26} & \heatcell{metricblue}{44}{78.57} & \heatcell{metricblue}{54}{77.91} & \heatcell{metricpurple}{35}{61.00} & \heatcell{metricpurple}{41}{68.93} & \heatcell{metricpurple}{39}{64.78} \\
\methodchip{4D9CAE}MSGSA~\cite{Lin2024MultiStageGeometricSemantic}           & \heatcell{metricblue}{35}{60.43} & \heatcell{metricblue}{60}{\second{89.01}} & \heatcell{metricblue}{46}{71.98} & \heatcell{metricpurple}{17}{47.99} & \heatcell{metricpurple}{60}{84.32} & \heatcell{metricpurple}{34}{61.22} \\
\methodchip{C29436}GCTNet~\cite{Guo2024GraphContextTransformation}           & \heatcell{metricblue}{55}{77.00} & \heatcell{metricblue}{45}{79.02} & \heatcell{metricblue}{55}{78.00} & \heatcell{metricpurple}{36}{61.12} & \heatcell{metricpurple}{41}{69.34} & \heatcell{metricpurple}{38}{64.31} \\
\methodchip{9372B2}BCLNet~\cite{Miao2024BCLNet}                              & \heatcell{metricblue}{57}{78.49} & \heatcell{metricblue}{50}{82.56} & \heatcell{metricblue}{58}{80.10} & \heatcell{metricpurple}{58}{77.39} & \heatcell{metricpurple}{55}{79.77} & \heatcell{metricpurple}{58}{78.31} \\
\baselineband \methodchip{476FB0}MatchMamba~\cite{Wu2025MatchMamba}                        & \heatcell{metricblue}{60}{\second{80.53}} & \heatcell{metricblue}{53}{84.27} & \heatcell{metricblue}{61}{\second{82.36}} & \heatcell{metricpurple}{60}{\second{78.27}} & \heatcell{metricpurple}{55}{80.04} & \heatcell{metricpurple}{59}{\second{79.15}} \\
\midrule
\ours \methodchip{C6384A}\oursname{SFMambaNet (ours)}                                   & \heatcell{metricblue}{62}{\best{82.49}} & \heatcell{metricblue}{54}{85.37} & \heatcell{metricblue}{62}{\best{83.16}} & \heatcell{metricpurple}{62}{\best{80.05}} & \heatcell{metricpurple}{59}{83.49} & \heatcell{metricpurple}{62}{\best{81.51}} \\
\bottomrule
\end{tabular}
\endgroup
\end{table}

\section{Homography estimation margins}
\label{sec:s_homography}

\begin{figure}[t]
  \centering
  \includegraphics[width=\linewidth]{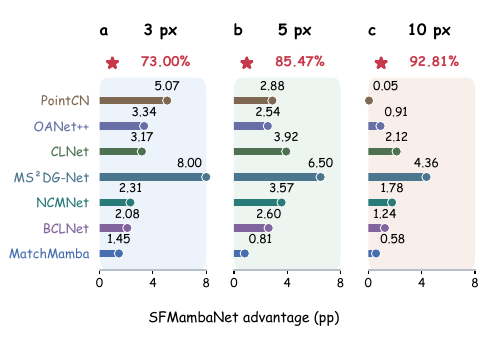}
  \caption{\textbf{Homography estimation on HPatches.} Accuracy gain of SFMambaNet over each baseline at corner-error thresholds of 3, 5, and 10 pixels, in percentage points. Positive values favor SFMambaNet; red stars show its absolute accuracies. All absolute accuracies are reported in Fig.~\ref*{fig:homography_bars} of the main paper.}
  \label{fig:s_homography}
\end{figure}

\Cref{fig:s_homography} plots the accuracy gains of SFMambaNet over the baselines in Fig.~\ref*{fig:homography_bars} of the main paper at each error threshold.
SFMambaNet outperforms all baselines at every threshold, and its margin over MatchMamba is largest at the strictest threshold of 3 pixels, where precise rather than merely successful estimates are required.

\section{Additional qualitative results}
\label{sec:s_qual}

\begin{figure*}[t]
  \centering
  \setlength{\tabcolsep}{0pt}
  \begin{tabular}{@{}p{0.248\textwidth}p{0.248\textwidth}p{0.248\textwidth}p{0.248\textwidth}@{}}
  \centering\footnotesize Putative correspondences & \centering\footnotesize BCLNet~\cite{Miao2024BCLNet} & \centering\footnotesize MatchMamba~\cite{Wu2025MatchMamba} & \centering\arraybackslash\footnotesize SFMambaNet (ours)
  \end{tabular}\\[1pt]
  \includegraphics[width=\textwidth]{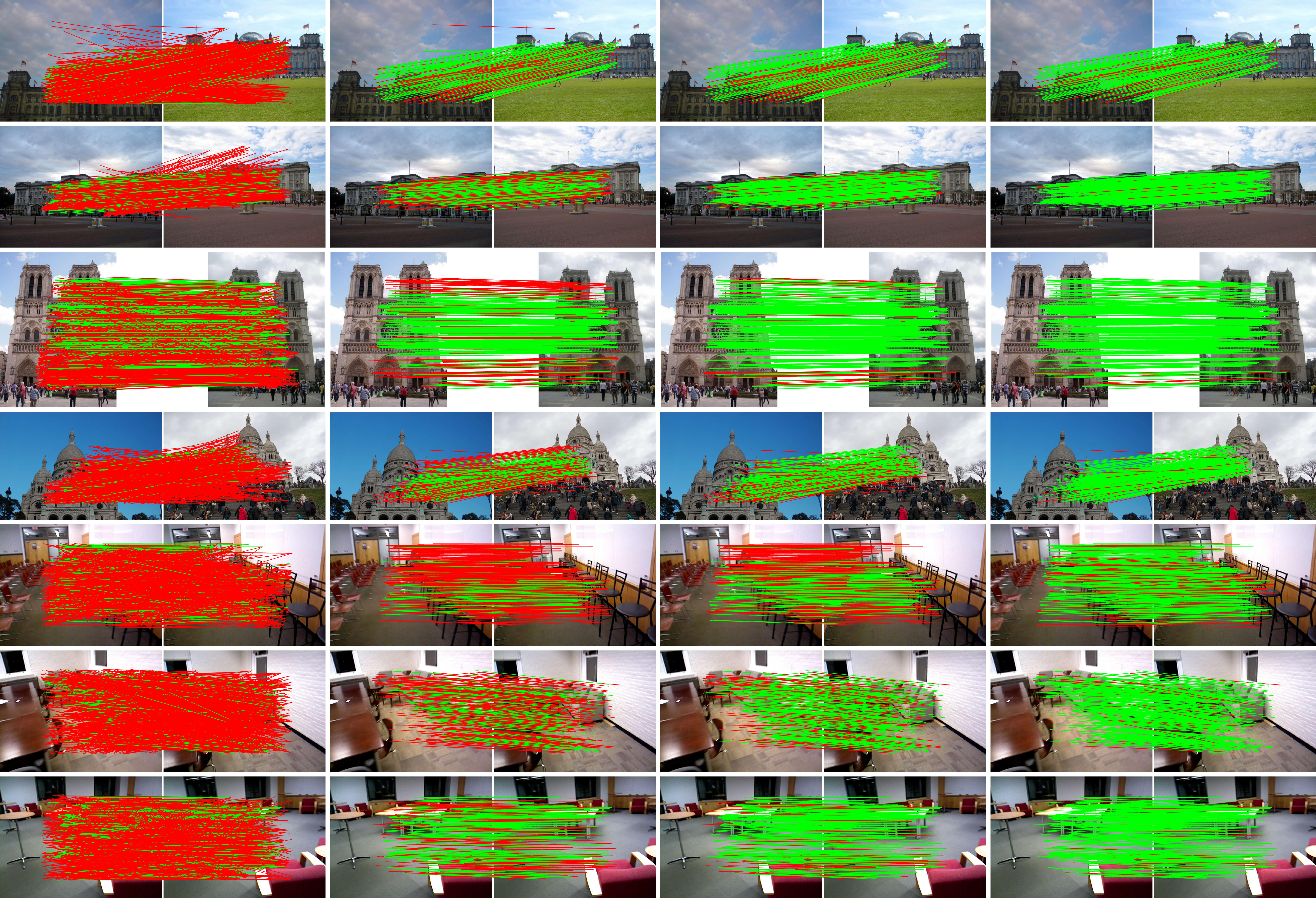}
  \caption{\textbf{Outlier removal on the unknown splits.} The first four rows are taken from YFCC100M and the last three from SUN3D. The pairs involve perspective changes, occlusions, repeated patterns, and textureless regions. Green and red lines denote inliers and outliers among the retained correspondences.}
  \label{fig:s_qual}
\end{figure*}

\Cref{fig:s_qual} shows all seven pairs from which Fig.~\ref*{fig:qual} of the main paper is selected.
On the outdoor pairs, BCLNet and MatchMamba retain outliers that run against the dominant motion, especially around repeated facade elements, while SFMambaNet keeps a denser set of inliers.
On the indoor pairs, where weak texture and repeated furniture make local appearance ambiguous, the differences are larger: the competitors retain many outliers in the lower part of the images, whereas SFMambaNet keeps mostly correct correspondences.

\section{Further discussion of the ablations}
\label{sec:s_ablation}

\paragraph{Controlled attribution.}
The results in Tab.~\ref*{tab:abl_main}, Fig.~\ref*{fig:gate_control}, and Tab.~\ref*{tab:encoding_control} of the main paper isolate the contributions of the components, frequency-domain gates, and geometric encoding.
The gate comparison preserves the cluster-ordering rule, both Mamba scans, residual fusion, and normalization; the encoding comparison preserves the attention operation while changing only its geometric input.
The experimental protocols and parameter counts are specified in \cref{sec:s_controlled_definitions}.

\paragraph{Executable component sequence.}
The cumulative variants in \cref{fig:lsga_progress} follow the sequence defined in \cref{sec:s_controlled_definitions}.
DiffPool is introduced jointly with its point-to-cluster interaction, followed by the ordered point-level BiMamba and then BiCSM.
Each step therefore uses only components introduced before it.
The Fourier encoding and local attention are treated as a joint SGEA addition in the cumulative plot; their separate contributions are examined in Tab.~\ref*{tab:encoding_control}.

\paragraph{Order sensitivity.}
\Cref{tab:input_permutations} tests the trained models directly: each checkpoint is run on randomly permuted inputs, and its predictions are compared with the unpermuted ones after restoring correspondence identities.
These end-to-end measurements complement the bounds of \cref{sec:s_order}.
The local intervention protocol is specified in \cref{sec:s_local_permutations}.

\paragraph{Role of the cluster order.}
The gates act along the cluster-induced order, and several results show that this order is what gives their frequencies a meaning.
On the raw coordinates of \cref{sec:s_teaser}, a cluster-wise order raises the share of inlier energy in the lowest tenth of the frequencies from 9.9\% to 84.2\%, while the outlier share rises only to 15.4\%.
Inside the trained network, along the learned order, the input of the first gate carries 2.19 times as much high-band energy per outlier position as per inlier position, and 1.61 times as much low-band energy per inlier position (\cref{sec:s_spectrum}).
Even the fixed low-pass control, which only smooths neighboring positions of this order, improves on identity gates by 1.55 points on unknown YFCC100M and 0.53 points on unknown SUN3D (Fig.~\ref*{fig:gate_control} of the main paper), so neighboring positions carry related information.
Finally, \cref{sec:s_order} shows that the order turns any reordering inside the clusters into high-frequency variation, which the learned gates attenuate.

\section{Isolated-gate frequency-domain diagnostic}
\label{sec:s_spectrum}

This diagnostic opens the first learned gate of the trained models and leads to three observations.
The gate keeps most of the low-frequency energy of its input and removes most of the high-frequency energy.
Per position, the high-frequency energy it receives is about twice as large at outliers as at inliers.
Restoring only its high-frequency gains costs more than four fifths of the AP lost by removing the gate, whereas matching the output energy changes AP by only $0.06$ points.

\begin{figure*}[t]
  \centering
  \includegraphics[width=\textwidth]{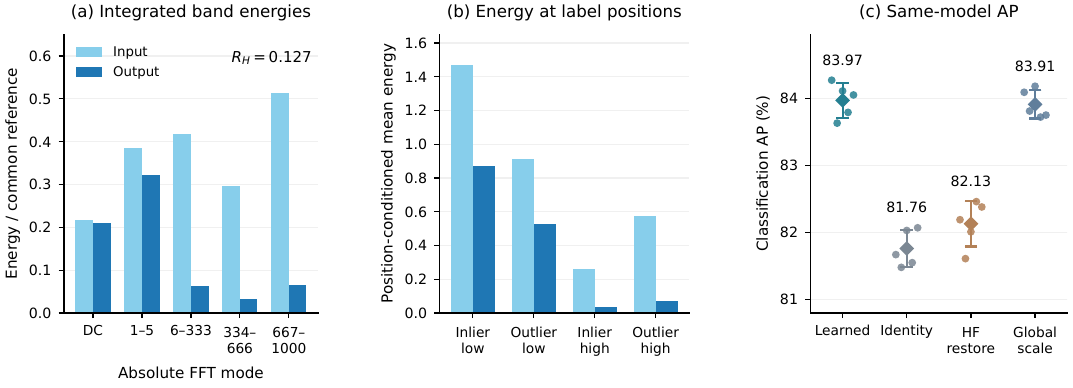}
  \caption{\textbf{An isolated forward-gate diagnostic.} (a) Integrated input and output FFT energies in five disjoint frequency groups, with DC as its own group, in units of a common checkpoint-specific reference. (b) Mean reconstructed low- and high-band feature energies at inlier and outlier positions. (c) First-stage, pre-top-$K$ AP under four same-model gate conditions. Dots represent five training seeds; error bars are sample standard deviations.}
  \label{fig:s_spectrum}
\end{figure*}

\subsection{Cohort and energy accounting}
\paragraph{Evaluation cohort and insertion point.}
We evaluate five trained models from the paired-seed experiment on the same cohort of $1000$ image pairs, with $250$ pairs from each of the four held-out YFCC100M test sequences.
Each pair has $N=2000$ candidate correspondences and $D=128$ feature channels.
The cohort therefore contains $2{,}000{,}000$ correspondence instances, comprising $397{,}843$ inliers and $1{,}602{,}157$ outliers, an inlier prevalence of $p=0.199$.
Inliers follow the paper's labeling convention: a symmetric epipolar distance below $10^{-4}$.
All $1000$ pairs contain both classes.
The diagnostic location is the forward gate in the first SIGM module of the first pruning stage: $X$ is its forward-Mamba output and $Y=T_GX$ denotes the output immediately after that gate, before the backward scan, residual fusion, or layer normalization.
For every comparison, the model weights are unchanged and the gate receives the same inputs.

\paragraph{Energy accounting and frequency groups.}
We compute diagnostic values using the unitary DFT along the correspondence axis, with
\begin{equation}
 E(Z;B)=\frac{1}{ND}\sum_{d=1}^{D}\sum_{\omega\in B}
       |\widehat Z_d(\omega)|^2.
 \label{eq:s4_energy}
\end{equation}
For real-valued features, a one-sided rFFT counts interior frequencies twice and DC and Nyquist once.
With $f=2|\omega|/N$, the bands are $L:f<1/3$, $M:1/3\le f<2/3$, and $H:f\ge2/3$.
For $N=2000$ these contain integer modes $0$--$333$, $334$--$666$, and $667$--$1000$ on the nonnegative side, respectively, and together cover all frequencies.
DC and modes $1$--$5$ are reported separately inside $L$.

\paragraph{Common energy reference.}
We express all energies in units of one positive reference $C_s$ per checkpoint $s$, the mean total gate-input energy on the same predetermined validation set.
The same $C_s$ is used for $X$, every gate output, all classes, and all bands of that checkpoint, and is determined without using unknown-split labels or selecting a gate condition.
The reference cancels in every before--after ratio, and cohort retention ratios use the pooled energy sums with identical weights for input and output.
\Cref{tab:s4_energy} reports the energy values in these units.

\begin{table}[t]
\centering
\caption{\textbf{Integrated FFT energies of the isolated gate} in common reference units. The first five rows partition the spectrum, and the last three summarize them. Retention is output over input energy.}
\label{tab:s4_energy}
\setlength{\tabcolsep}{4pt}
\small
\renewcommand{\arraystretch}{1.08}
\begin{tabular}{@{}cccc@{}}
\toprule
\neutralheader{Frequency group} & \metricheader{Input $X$} & \auxheader{Output $Y$} & \altmetricheader{Retention} \\
\midrule
\tabbaseline DC, $\omega=0$                 & 0.217 & 0.209 & 0.965 \\
$1\le|\omega|\le5$            & 0.386 & 0.323 & 0.837 \\
\tabbaseline $6\le|\omega|\le333$          & 0.419 & 0.062 & 0.148 \\
$334\le|\omega|\le666$        & 0.297 & 0.034 & 0.113 \\
\tabbaseline $667\le|\omega|\le1000$       & 0.513 & 0.065 & 0.127 \\
\midrule
\tabstablerow Low band $L$                  & 1.022 & 0.594 & 0.582 \\
\tabstablerow Low band excluding DC         & 0.805 & 0.385 & 0.478 \\
\tabstablerow All frequencies               & 1.832 & 0.693 & 0.378 \\
\bottomrule
\end{tabular}
\end{table}

\paragraph{Relation to the gain bounds.}
The values in \cref{tab:s4_energy} respect the first-stage forward-gate bound $G_{\max}=1.15$ and $\varepsilon=0.40$ beyond $|\omega|=5$ in \cref{tab:s_gate_response}.
For the original gate, this implies $E(Y;H)/E(X;H)\le0.16$, and the same bound applies to $M$, since both bands lie entirely beyond mode five; the measured retentions are $0.127$ and $0.113$.
The low band $L$ is much wider than the cutoff $f_c=2\cdot5/2000=0.005$ of the order argument: $59.0\%$ of its input energy lies at $|\omega|\le5$, including a $21.2\%$ DC share, and its retention after removing DC is $0.478$.
The gate response behind these values is interpolated from its $32$ complex control coefficients.

\subsection{Class-conditioned energies}
\paragraph{Position-conditioned class energies.}
For the class diagnostics, we first reconstruct $Z_b=\mathcal{F}^{-1}(\mathbf{1}_b\widehat Z)$ for each full sequence and each band $b$, and then average $|Z_b(i,d)|^2$ over channels and correspondence positions carrying a ground-truth label $c$.
Across the cohort, we divide the total energy at those positions by the total number of class-$c$ positions and channels, so the energies describe the mixed sequence that the network actually processes.
The same reference $C_s$ is used for both classes, and in every band the class means satisfy $E(Z;b)=pE_{\mathrm{in},b}+(1-p)E_{\mathrm{out},b}$.
At the gate input, an inlier position carries $1.61$ times the low-band energy of an outlier position ($1.467$ versus $0.911$), whereas an outlier position carries $2.19$ times the high-band energy of an inlier position ($0.575$ versus $0.263$); the mid-band energies are similar ($0.312$ and $0.293$).
The gate reduces these energies to $0.870$ and $0.526$ in the low band, $0.035$ and $0.073$ in the high band, and $0.034$ and $0.033$ in the mid band.
It keeps a similar fraction of each band at both classes (low band: $0.593$ and $0.577$; high band: $0.134$ and $0.126$), so the high-band energy it removes lies mainly at outlier positions, where that energy is concentrated.
We compare these energies band by band, since bands restricted to the positions of one class are not orthogonal.

\subsection{Same-model gate interventions}
\paragraph{Four same-model gate conditions.}
The interventions modify only the selected forward gate, leave its input and every model weight unchanged, and recompute its complete downstream path to the first-stage classification head.
The four outputs are: the original learned response $Y_G=T_GX$; identity $Y_I=X$; high-frequency gain restoration $Y_R=T_{\widetilde G}X$; and a global-scale control $Y_C=cY_G$.
On $H$ only, restoration uses
\begingroup\small
\setlength{\thickmuskip}{4mu}
\begin{equation}
 \widetilde G(\omega,d)=
 \begin{cases}
 \max\{|G(\omega,d)|,1\}\dfrac{G(\omega,d)}{|G(\omega,d)|},
 & |G(\omega,d)|>10^{-8},\\[3pt]
 G(\omega,d),& |G(\omega,d)|\le10^{-8}.
 \end{cases}
 \label{eq:s4_restore}
\end{equation}
\endgroup
Outside $H$, we keep $\widetilde G=G$.
The effective conjugate-symmetric response preserves the original phase wherever restoration is applied, and the phase at zero or tiny gain is left unchanged; no high-band input energy falls into such tiny-gain bins.
Because the original high-band gains are below one, restoration gives $E(Y_R;H)/E(X;H)=1$.

\paragraph{Pair-specific scale matching.}
Scale matching uses $c=\|Y_R\|_F/\|Y_G\|_F$ separately for each image pair, without class labels, so that $Y_C$ and $Y_R$ have identical total energy while $Y_C$ retains the original output's relative spectral shape and phase.
For an input with the energies in \cref{tab:s4_energy}, restoration raises the total output energy from $0.693$ to $0.594+0.034+0.513=1.141$, which gives $c=\sqrt{1.141/0.693}=1.283$ and scaled $L/M/H$ energies of $0.979/0.055/0.107$.

\paragraph{AP before the first pruning decision.}
Average precision (AP) uses the first-stage classification scores for all $2000$ input correspondences before top-$K$ selection.
The later SIGM modules and the classification head process each intervention's features as usual, so AP reflects the full downstream effect of the change.
We compute AP separately for each image pair, average pairs within each test sequence, and give the four sequence means equal weight for each seed.
Labels, scoring conventions, and sequence weights are identical across the four conditions, and \cref{tab:s4_ap} gives five-seed values and per-sequence means.

\begin{table*}[t]
\centering
\caption{\textbf{First-stage pre-top-$K$ AP under four same-model gate conditions.} AP is in \%. Seed rows give each checkpoint's equally weighted mean over four sequence means; sequence rows average that sequence's pair-mean AP over the five checkpoints. The final row gives the five-seed mean and sample standard deviation. Bold marks the row-wise highest AP.}
\label{tab:s4_ap}
\setlength{\tabcolsep}{11pt}
\small
\renewcommand{\arraystretch}{1.08}
\begin{tabular}{@{}ccccc@{}}
\toprule
\auxheader{Summary} & \metricheader{Original $G$} & \neutralheader{Identity} & \warmheader{HF restoration} & \altmetricheader{Global scale} \\
\midrule
\tabbaseline Seed 1 & \tabbest{84.27} & 81.67 & 82.19 & 84.09 \\
Seed 2 & 83.63 & 81.48 & 81.61 & \tabbest{83.81} \\
\tabbaseline Seed 3 & 84.11 & 82.03 & 82.01 & \tabbest{84.18} \\
Seed 4 & \tabbest{83.79} & 81.55 & 82.46 & 83.72 \\
\tabbaseline Seed 5 & \tabbest{84.05} & 82.07 & 82.38 & 83.75 \\
\midrule
\tabstablerow Sequence 1 & 86.93 & 85.14 & 85.68 & \tabbest{87.01} \\
Sequence 2 & \tabbest{84.42} & 82.31 & 82.79 & 84.17 \\
\tabstablerow Sequence 3 & 82.67 & 80.46 & 81.08 & \tabbest{82.84} \\
Sequence 4 & \tabbest{81.86} & 79.13 & 78.97 & 81.62 \\
\midrule
\tabfinal Mean $\textcolor{tablesd}{\pm}$ seed SD & $\tabbest{83.97}\tabunc{0.26}$ & $81.76\tabunc{0.27}$ & $82.13\tabunc{0.34}$ & $83.91\tabunc{0.21}$ \\
\bottomrule
\end{tabular}
\end{table*}

\paragraph{Paired seed uncertainty.}
Paired contrasts use the five seed-level differences after the within-sequence pair means and equal sequence weighting have been computed.
For differences $d_s$, we report $\bar d\pm s_d$ and the paired $95\%$ interval $\bar d\pm t_{4,0.975}s_d/\sqrt5$, where $t_{4,0.975}=2.776$ and $s_d$ is the sample standard deviation.
\Cref{tab:s4_paired_ci} gives the resulting intervals in pp; they describe the variability across the five training seeds.

\begin{table}[t]
\centering
\caption{\textbf{Paired AP differences across five training seeds}, in pp. Intervals use the paired Student-$t$ construction with four degrees of freedom.}
\label{tab:s4_paired_ci}
\setlength{\tabcolsep}{3.5pt}
\small
\renewcommand{\arraystretch}{1.08}
\renewcommand{\tabularxcolumn}[1]{m{#1}}
\begin{tabularx}{\linewidth}{@{}>{\centering\arraybackslash}X >{\centering\arraybackslash}m{0.245\linewidth} >{\centering\arraybackslash}m{0.29\linewidth}@{}}
\toprule
\metricheader{Paired contrast} & \warmheader{Mean $\textcolor{tablesd}{\pm}$ SD} & \auxheader{\makecell[c]{$95\%$\\paired interval}} \\
\midrule
\tabbaseline Original $G$ $-$ Identity       & $2.21\tabunc{0.24}$ & $\textcolor{secondc}{[1.91,\ 2.51]}$ \\
Original $G$ $-$ HF restoration & $1.84\tabunc{0.33}$ & $\textcolor{secondc}{[1.43,\ 2.25]}$ \\
\rowcolor{tableneutral} Original $G$ $-$ Global scale   & $0.06\tabunc{0.19}$ & $\textcolor{tablesd}{[-0.18,\ 0.30]}$ \\
\tabstablerow Global scale $-$ HF restoration & $1.78\tabunc{0.44}$ & $\textcolor{secondc}{[1.23,\ 2.33]}$ \\
\bottomrule
\end{tabularx}
\end{table}

\paragraph{Interpretation.}
The trained gate acts as a low-pass filter: it keeps $58\%$ of the low-band input energy and $13\%$ of the high-band energy (\cref{tab:s4_energy}), and the high-band energy it removes is concentrated at outlier positions.
The network relies on this frequency shape rather than on the output energy: restoring the high-frequency gains costs $1.84$ of the $2.21$ AP points lost without the gate, whereas rescaling the original output to the same total energy changes AP by only $0.06$ points (\cref{tab:s4_paired_ci}).
Together with the capacity-matched and fixed-filter controls of \cref{sec:s_controlled_definitions}, which retrain complete models, these interventions place the benefit of the gate in its frequency response.
The diagnostic changes one gate at a time with all weights fixed; \cref{sec:s_local_permutations} complements it with tests of complete modules and of the complete network.

\section{Complete efficiency results}
\label{sec:s_efficiency}

\begin{figure}[t]
\centering
\includegraphics[width=\linewidth]{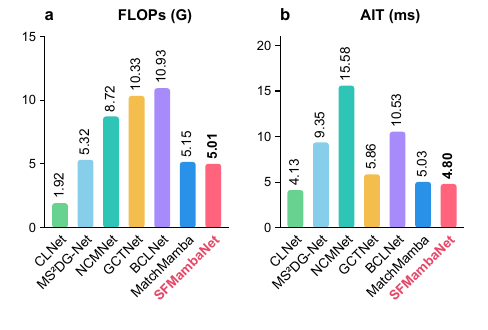}
\caption{\textbf{Computational cost} on YFCC100M Unknown with SIFT: FLOPs (G) and average inference time (AIT, ms); lower is better. Values are from \cref{tab:s_efficiency} and refer to complete models. Controlled-ablation budgets are reported separately.}
\label{fig:s_efficiency_costs}
\label{fig:s_efficiency_scatter}
\end{figure}

\Cref{tab:s_efficiency} reports all five efficiency metrics corresponding to the accuracy--cost comparison in Fig.~\ref*{fig:efficiency_costs} of the main paper.
\Cref{fig:s_efficiency_costs} gives the exact FLOPs and inference runtimes.

\begin{table}[!htbp]
\centering
\caption{\textbf{Efficiency} on the unknown split of YFCC100M with SIFT: \mapfive (\%), parameters (M), FLOPs (G), and average training (ART) and inference (AIT) runtimes (ms). Baseline values are those reported in~\cite{Wu2025MatchMamba}. Darker cells indicate better values within each column.}
\label{tab:s_efficiency}
\setlength{\tabcolsep}{3pt}
\footnotesize
\begingroup
\renewcommand{\arraystretch}{1.05}
\renewcommand{\best}[1]{\textcolor[HTML]{7A1930}{\textbf{#1}}}
\renewcommand{\second}[1]{\textcolor{secondc}{\underline{#1}}}
\begin{tabular}{@{}c ccccc@{}}
\toprule
\multicolumn{1}{@{}c}{\multirow{2}{*}[-\dimexpr(\aboverulesep+\belowrulesep+\cmidrulewidth)/2\relax]{Method}} & \metricheader{Accuracy} & \multicolumn{2}{c}{\auxheader{Model cost}} & \multicolumn{2}{c}{\altmetricheader{Runtime}} \\
\cmidrule(lr){2-2}\cmidrule(lr){3-4}\cmidrule(lr){5-6}
 & \metricheader{\mapfive$\uparrow$} & \metricheader{Params$\downarrow$} & \auxheader{FLOPs$\downarrow$} & \altmetricheader{ART$\downarrow$} & \warmheader{AIT$\downarrow$} \\
\midrule
\methodchip{72A877}CLNet~\cite{Zhao2021ProgressiveCorrespondencePruning} & \heatcell{bestc}{20}{54.05} & \heatcell{metricblue}{54}{\best{0.95}} & \heatcell{metricteal}{54}{\best{1.92}} & \heatcell{metricpurple}{54}{\best{181.87}} & \heatcell{metricgold}{54}{\best{4.13}} \\
\methodchip{6198B7}MS$^2$DG-Net~\cite{Dai2022MS2DGNet} & \heatcell{bestc}{12}{49.13} & \heatcell{metricblue}{42}{2.53} & \heatcell{metricteal}{38}{5.32} & \heatcell{metricpurple}{41}{396.19} & \heatcell{metricgold}{35}{9.35} \\
\methodchip{2F8C8A}NCMNet~\cite{Liu2023ProgressiveNeighborConsistencyMining} & \heatcell{bestc}{36}{63.52} & \heatcell{metricblue}{28}{4.49} & \heatcell{metricteal}{22}{8.72} & \heatcell{metricpurple}{12}{855.14} & \heatcell{metricgold}{12}{15.58} \\
\methodchip{C29436}GCTNet~\cite{Guo2024GraphContextTransformation} & \heatcell{bestc}{37}{63.80} & \heatcell{metricblue}{31}{4.09} & \heatcell{metricteal}{15}{10.33} & \heatcell{metricpurple}{39}{428.80} & \heatcell{metricgold}{48}{5.86} \\
\methodchip{9372B2}BCLNet~\cite{Miao2024BCLNet} & \heatcell{bestc}{41}{66.08} & \heatcell{metricblue}{12}{6.65} & \heatcell{metricteal}{12}{10.93} & \heatcell{metricpurple}{22}{687.15} & \heatcell{metricgold}{31}{10.53} \\
\baselineband \methodchip{476FB0}MatchMamba~\cite{Wu2025MatchMamba} & \heatcell{bestc}{48}{\second{70.23}} & \heatcell{metricblue}{46}{2.08} & \heatcell{metricteal}{39}{5.15} & \heatcell{metricpurple}{41}{383.31} & \heatcell{metricgold}{51}{5.03} \\
\midrule
\ours \methodchip{C6384A}\oursname{SFMambaNet (ours)} & \heatcell{bestc}{54}{\best{73.83}} & \heatcell{metricblue}{46}{\second{2.01}} & \heatcell{metricteal}{40}{\second{5.01}} & \heatcell{metricpurple}{42}{\second{370.42}} & \heatcell{metricgold}{52}{\second{4.80}} \\
\bottomrule
\end{tabular}
\endgroup
\end{table}

\section{Details of the first-page figure}
\label{sec:s_teaser}

Fig.~\ref*{fig:teaser} of the main paper demonstrates the premise of Sec.~\ref*{sec:sigm} on the raw coordinates of one image pair: a set of matches has no spectrum until it is given an order, and a cluster-wise order separates the two classes.

\paragraph{Image pair.}
The two photographs show the south facade of Notre-Dame de Paris.
They were taken from Wikimedia Commons, where their author, Chabe01, released them under CC0 (``Cath\'edrale Notre-Dame fa\c{c}ade sud Paris~5'' and ``Paris~3'').
We use the renditions that are 1920 pixels wide and keep their top 1140 rows, which removes the passers-by at street level.
Panel~(a) displays them slightly desaturated, and panels~(b) and~(c) show the first one in gray; all computations use the original pixels.

\paragraph{Matches and labels.}
As for the benchmarks, we extract 2000 SIFT keypoints per image and match every keypoint of the first image to its nearest neighbor in the second, which gives $N=2000$ putative matches.
The labels follow an estimated epipolar geometry: a fundamental matrix is fitted with MAGSAC++~\cite{Barath2020MAGSACpp} (threshold of 1 pixel) to the 377 mutual nearest neighbors that pass a ratio test at 0.8, and a match is labeled as an inlier if its distance to the epipolar line is below 2 pixels in both images.
This yields 462 inliers and 1538 outliers; between 428 and 499 matches are inliers at thresholds between 1 and 5 pixels.
To keep the photographs visible, panel~(a) draws only a random subset of the matches, whose size is stated in the figure.

\paragraph{Scan orders.}
The random order of panel~(b) is a uniformly random permutation of the matches.
For the cluster-wise order of panel~(c), $k$-means groups the keypoints of the first image into 32 clusters by position (14 to 156 matches per cluster), a partition that involves no learned parameters; the clusters are visited one after another, and the matches of a cluster in random order.
This mirrors the cluster-induced order $\Pi$ of Eq.~(\ref*{eq:order}), whose order inside a block is arbitrary as well.
On the left of both panels, the color of a keypoint encodes its position in the scan; the bar above the sequence in the middle repeats the colors, and its gaps mark the cluster boundaries.

\paragraph{Sequences and spectra.}
The sequence in the middle of panels~(b) and~(c) is the horizontal coordinate $u'$ at which each match lands in the second image (cf.\ Eq.~(\ref*{eq:input}); here in units of the image width), in scan order; dots are outliers, and the line connects consecutive inliers.
For either class, let $s\in\mathbb{R}^{n}$ contain the values of $u'$ of its $n$ matches in scan order with their mean removed, and let $\hat s$ be its discrete Fourier transform.
The right of each panel shows the relative spectrum
\begin{equation}
  P\Big(\frac{k}{n}\Big)=\frac{\big|\hat s(k)\big|^2}{\frac{1}{K}\sum_{j=1}^{K}\big|\hat s(j)\big|^2},\qquad k=1,\dots,K=\lfloor n/2\rfloor,
  \label{eq:s_teaser}
\end{equation}
of both classes; its expectation is one at every frequency when the order is random (dotted line).
We average $P$ over 200 random orders in panel~(b) and over 200 random orders inside the clusters in panel~(c).
The plotted curves are averaged in 16 logarithmically spaced frequency bins between the lowest frequency of the inlier sequence and the Nyquist frequency.
Frequencies are given in cycles per match of the class.

\paragraph{Effect of scan order.}
In the cluster-wise order of panel~(c), the lowest tenth of the frequencies holds 84.2\% of the energy of the inlier sequence and 15.4\% of that of the outlier sequence, compared with 9.9\% and 10.0\% in the random order of panel~(b); with both coordinates $(u',v')$, the shares are 83.9\% and 16.5\%.
The flat spectra of panel~(b) follow from the flat expected periodogram of a randomly permuted sequence.
The observation is panel~(c), in which the same matches separate once neighboring matches are scanned together.
The small rise of the outlier share, from 10.0\% to 15.4\%, comes from the few mismatches of neighboring keypoints that happen to agree.
Inside the trained network, the same contrast appears on the mixed sequence along the learned order: at the input of the first gate, an outlier position carries 2.19 times the high-band energy of an inlier position, and an inlier position 1.61 times the low-band energy of an outlier position (\cref{sec:s_spectrum}).
This separation motivates the design of SIGM. The model and gate comparisons appear in Tab.~\ref*{tab:abl_main} and Fig.~\ref*{fig:gate_control} of the main paper and \cref{fig:s_spectrum}.

\section{Controlled comparison definitions}
\label{sec:s_controlled_definitions}

\subsection{Protocol and gate controls}
\paragraph{Evaluation conditions.}
This section and the following sections report controlled comparisons and diagnostics of the trained models.
Every configuration uses five paired training seeds and the same data splits, training budget, checkpoint selection rule, and candidate correspondence sets.
The numerical comparisons use SIFT on the unknown splits of YFCC100M and SUN3D, report pose \mapfive without RANSAC, and express differences in percentage points (pp).
The reported means and sample standard deviations are computed across the five independent training runs.

\paragraph{Capacity-matched gate controls.}
The two pruning stages each contain three SIGM modules with two gates per module, giving twelve gates in total.
Each gate stores $128\times32$ complex coefficients, or $8192$ real trainable scalars; interpolation changes the response length but not the parameter count.
The twelve gates therefore contain $98{,}304$ real trainable scalars.
The shared full-Fourier trunk excluding these gates has $1{,}911{,}808$ trainable parameters, giving a complete-model count of $2{,}010{,}112$.
The complex control coefficients affect the interior frequency response; self-conjugate bins have real values and the effective response obeys conjugate symmetry.

For the pointwise capacity control, we replace each gate with a bias-free $128\to32\to128$ MLP with a parameter-free activation.
Each replacement has $128\cdot32+32\cdot128=8192$ parameters, giving the same $98{,}304$-parameter budget over twelve locations.
Its operation is $T_{\mathrm{MLP}}(x_i)=x_i+W_2\operatorname{GELU}(W_1x_i)$, with nonzero $W_1$ and zero $W_2$ at initialization, and no additional normalization.
The scan skeleton, insertion locations, residual fusion, normalization, assignment-network architecture, and sorting rule remain the same architectural choices in all gate variants; weights are trained independently for each configuration.
This control asks whether the same number of point-wise parameters can buy the improvement.

\paragraph{Fixed Gaussian low-pass control.}
The fixed filters use the real, zero-phase response
\begin{equation}
 H_{\tau}(f)=\exp\!\left[-\ln(2)\left(\frac{f}{\tau}\right)^2\right],
 \qquad f=\frac{2|\omega|}{N}.
 \label{eq:s_fixed_gaussian}
\end{equation}
The bandwidth candidates are $\tau\in\{0.01,0.025,0.05,0.10,0.20\}$.
For each dataset, the forward and backward gate positions use validation-selected bandwidths; each selected value is shared across stages, SIGM layers, and the five paired seeds.
The bandwidths are locked before unknown-split evaluation.
Identity gates use the constant response $G(\omega)=1$.
Both controls have no trainable gate parameters.
Multiplying the spectrum by the real, even response of \cref{eq:s_fixed_gaussian} smooths the cluster-ordered sequence with a fixed, Gaussian-shaped kernel, \ie, a circular convolution along the order.
The controls therefore bracket the learned gates: the point-wise MLP matches their capacity without mixing positions, and the fixed filter mixes neighboring positions without learned frequency selectivity.
The learned gates outperform both on both datasets (Fig.~\ref*{fig:gate_control} of the main paper).

\subsection{Encoding controls}
\paragraph{Capacity-matched geometric encoders.}
The local controls preserve the same LGA queries, keys, values, neighborhood, attention normalization, and fusion, and change only the geometric embedding.
The no-explicit-geometry control sets $E_{\mathit{geo}}=0$ and removes its encoder.
The raw-coordinate control maps $\Delta C\in\mathbb{R}^4$ through a biased $4\to247\to128$ MLP.
Its parameter count per stage is $4\cdot247+247+247\cdot128+128=32{,}979$.
The Fourier control uses the fixed $4\to128$ sine/cosine feature map followed by a biased $128\to128\to128$ MLP, with $33{,}024$ parameters per stage.
Across two stages, the respective encoder budgets are $65{,}958$ and $66{,}048$, a difference of only $90$ parameters; the fixed Gaussian projection is not trainable.
Both encoders use GELU as the parameter-free activation.
The fixed-lift control uses a fixed $A\in\mathbb{R}^{128\times4}$ followed by the same $128\to128\to128$ MLP as the Fourier encoder, with identical trainable capacity.
A scalar fixed from training inputs matches its input RMS to the Fourier feature RMS; it is locked before validation and testing.
Neither $A$ nor this scalar is trainable.

\begin{table}[t]
\centering
\caption{\textbf{Parameter counts for the controlled comparisons.} Counts are real trainable scalars. The common full-Fourier trunk excluding the twelve gates has $1{,}911{,}808$ parameters. The encoder totals cover both pruning stages.}
\label{tab:s_controlled_budgets}
\setlength{\tabcolsep}{3pt}
\renewcommand{\arraystretch}{1.10}
\small
\renewcommand{\tabularxcolumn}[1]{m{#1}}
\begin{tabularx}{\linewidth}{@{}>{\centering\arraybackslash}Xccc@{}}
\toprule
\neutralheader{Configuration} & \auxheader{Gate/control} & \metricheader{Encoder} & \costheader{Total} \\
\midrule
Fourier + identity & 0 & 66,048 & 1,911,808 \\
Fourier + fixed low-pass & 0 & 66,048 & 1,911,808 \\
Fourier + one learned position & 49,152 & 66,048 & 1,960,960 \\
\tabbaseline
Fourier + pointwise MLP & 98,304 & 66,048 & 2,010,112 \\
\tabfinal
\oursname{Fourier + learned gates} & 98,304 & 66,048 & 2,010,112 \\
\tabbaseline
Fixed linear lift + learned gates & 98,304 & 66,048 & 2,010,112 \\
\tabbaseline
Raw coordinates + learned gates & 98,304 & 65,958 & 2,010,022 \\
No explicit geometry + learned gates & 98,304 & 0 & 1,944,064 \\
\bottomrule
\end{tabularx}
\end{table}

\subsection{Cumulative LSGA control}
\paragraph{Dependency-respecting LSGA sequence.}
The cumulative variants in \cref{fig:abl_lsga} contain no SIGM.
The baseline omits LSGA; the SGEA step adds the geometric encoder together with LGA, and Tab.~\ref*{tab:encoding_control} of the main paper separates the encoder from the attention.
DiffPool and multi-scale interaction (DP+MSI) are then introduced jointly: we compute $G^C,S$ from $F^M$, set $F^G=F^M$ and $F^L=G^C$, and apply the cross-scale fusion defined in the main paper.
The next variant computes $F^G=\Pi^{-1}\operatorname{BiMamba}(\Pi F^M)$ using the already available assignment $S$.
The final variant replaces $F^L=G^C$ with $F^L=\operatorname{BiCSM}(G^C)$.
Thus the sequence is baseline, SGEA, DP+MSI, ordered BiMamba, and BiCSM, and each step uses only components introduced before it.
Every added branch raises both metrics, and the final values match the LSGA-only row of Tab.~\ref*{tab:abl_main} of the main paper.

\begin{figure}[t]
\centering
\begin{tikzpicture}[x=1.60cm,y=.085cm]
\begin{scope}
  \clip (-.15,0) rectangle (4.15,35);
  \shade[top color=teal!4!white,bottom color=blue!3!white]
    (-.15,0) rectangle (4.15,35);
\end{scope}
\foreach \ordinate/\ticktext in {3/55,13/65,23/75,33/85} {
  \draw[gray!18] (-.15,\ordinate)--(4.15,\ordinate);
  \node[anchor=east,font=\scriptsize] at (-.20,\ordinate) {\ticktext};
}
\draw[gray!70] (-.15,35)--(-.15,0)--(4.15,0);
\node[rotate=90,font=\footnotesize] at (-.52,17.5) {mAP (\%)};
\foreach \position/\steptext in {0/Base,1/SGEA,2/{DP+MSI},3/{Ordered\\BiMamba},4/BiCSM} {
  \node[anchor=north,align=center,font=\scriptsize] at (\position,-1.8) {\steptext};
}
\begin{scope}
  \clip (-.15,0) rectangle (4.15,35);
  \draw[blue!65!black,line width=24pt,opacity=.07,line cap=round,line join=round]
    (0,3.87)--(1,7.27)--(2,9.03)--(3,11.08)--(4,12.71);
  \draw[blue!65!black,line width=16pt,opacity=.10,line cap=round,line join=round]
    (0,3.87)--(1,7.27)--(2,9.03)--(3,11.08)--(4,12.71);
\end{scope}
\draw[blue!65!black,line width=.9pt]
  (0,3.87)--(1,7.27)--(2,9.03)--(3,11.08)--(4,12.71);
\foreach \position/\ordinate/\deviation in {0/3.87/.48,1/7.27/.34,2/9.03/.52,3/11.08/.31,4/12.71/.39} {
  \draw[blue!65!black] (\position,{\ordinate-\deviation})--(\position,{\ordinate+\deviation});
  \draw[blue!65!black] ({\position-.025},{\ordinate-\deviation})--({\position+.025},{\ordinate-\deviation});
  \draw[blue!65!black] ({\position-.025},{\ordinate+\deviation})--({\position+.025},{\ordinate+\deviation});
  \fill[blue!65!black] (\position,\ordinate) circle[radius=1.6pt];
}
\begin{scope}
  \clip (-.15,0) rectangle (4.15,35);
  \draw[teal!75!black,line width=24pt,opacity=.07,line cap=round,line join=round]
    (0,23.13)--(1,26.19)--(2,28.02)--(3,29.94)--(4,31.32);
  \draw[teal!75!black,line width=16pt,opacity=.10,line cap=round,line join=round]
    (0,23.13)--(1,26.19)--(2,28.02)--(3,29.94)--(4,31.32);
\end{scope}
\draw[teal!75!black,line width=.9pt]
  (0,23.13)--(1,26.19)--(2,28.02)--(3,29.94)--(4,31.32);
\foreach \position/\ordinate/\deviation in {0/23.13/.31,1/26.19/.27,2/28.02/.38,3/29.94/.22,4/31.32/.26} {
  \draw[teal!75!black] (\position,{\ordinate-\deviation})--(\position,{\ordinate+\deviation});
  \draw[teal!75!black] ({\position-.025},{\ordinate-\deviation})--({\position+.025},{\ordinate-\deviation});
  \draw[teal!75!black] ({\position-.025},{\ordinate+\deviation})--({\position+.025},{\ordinate+\deviation});
  \node[rectangle,fill=teal!75!black,inner sep=1.5pt] at (\position,\ordinate) {};
}
\draw[blue!65!black,line width=.9pt] (.45,38)--(.75,38);
\fill[blue!65!black] (.60,38) circle[radius=1.6pt];
\node[anchor=west,font=\scriptsize] at (.80,38) {$5^\circ$};
\draw[teal!75!black,line width=.9pt] (2.25,38)--(2.55,38);
\node[rectangle,fill=teal!75!black,inner sep=1.5pt] at (2.40,38) {};
\node[anchor=west,font=\scriptsize] at (2.60,38) {$20^\circ$};
\end{tikzpicture}
\caption{Cumulative LSGA control on YFCC100M Unknown (five-seed mean $\pm$ SD). SIGM is disabled. DP and MSI are enabled jointly before ordered BiMamba, so clustering has an output path and the scan order exists. BiCSM completes LSGA. Error bars show SD.}
\label{fig:lsga_progress}
\label{fig:abl_lsga}
\end{figure}

\subsection{Crossed controls and gate placement}
\begin{figure}[t]
\centering
\includegraphics[width=\linewidth]{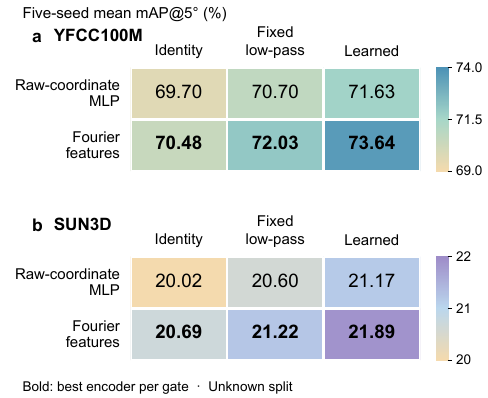}
\caption{\textbf{Crossed encoder and gate controls.} Colors and labels show five-seed mean pose \mapfive (\%) on YFCC100M (top) and SUN3D (bottom) unknown splits, with separately labeled color scales. All configurations share LGA and the bidirectional scan skeleton; only the encoder and gate type differ. Shared configurations reuse main-paper results. Bold marks the best value within each gate column.}
\label{fig:s_crossed_controls}
\label{tab:s_crossed_controls}
\end{figure}

\paragraph{Crossed gate--encoder control.}
Learned gates achieve higher accuracy with either encoder, and Fourier geometry improves accuracy with each gate type (\cref{fig:s_crossed_controls}), so each frequency component helps on its own.
The $2\times3$ factorial comparison in \cref{fig:s_crossed_controls} crosses raw-coordinate and Fourier geometric encoders with identity, fixed low-pass, and learned gates.
The LGA operation and the two-direction Mamba skeleton are held fixed as architecture definitions.

\paragraph{Gate placement.}
The results in \cref{tab:s_gate_positions} distinguish a gate after the forward scan from a gate after the backward scan.
In every row the backward scan consumes the forward branch produced by that row, and the residual connection and normalization remain unchanged.
Enabling one position introduces six gates, or $49{,}152$ parameters; enabling both introduces twelve gates.
The resulting single-position configuration has $1{,}960{,}960$ parameters.
The forward-only and backward-only comparisons isolate the effect of each insertion position without replacing a bidirectional module with a unidirectional one.

\begin{table}[t]
\centering
\caption{\textbf{Gate-placement controls.} I denotes identity and L a learned gate; these choices are applied consistently to the three SIGM modules in each pruning stage. Pose \mapfive is in \%, and parameters are in millions. Both Mamba directions are present in every row.}
\label{tab:s_gate_positions}
\setlength{\tabcolsep}{5pt}
\renewcommand{\arraystretch}{1.10}
\small
\begin{tabular}{@{}ccccc@{}}
\toprule
\neutralheader{Forward} & \neutralheader{Backward} & \costheader{Params (M)} & \metricheader{YFCC100M} & \altmetricheader{SUN3D} \\
\midrule
\tabbaseline
I & I & 1.912 & 70.48 & 20.69 \\
L & I & 1.961 & 72.85 & 21.52 \\
I & L & 1.961 & 71.95 & 21.18 \\
\tabfinal
\oursname{L} & \oursname{L} & 2.010 & \tabbest{73.64} & \tabbest{21.89} \\
\bottomrule
\end{tabular}
\end{table}

\section{Local permutation interventions}
\label{sec:s_local_permutations}

\subsection{Within-cluster interventions}
\paragraph{Checkpoint and module controls.}
We evaluate each of the five trained checkpoints in deterministic evaluation mode with all weights unchanged.
For each pruning stage, we intervene in one of its three SIGM modules at a time, leaving the other five SIGM modules unperturbed.
For this diagnostic, we write features as $X\in\mathbb{R}^{N\times D}$ with correspondences in rows, a notational transpose of the main paper's $D\times N$ layout, so $\Pi$ and $Q$ act by left multiplication on the correspondence axis.
At the selected module, we form $x=\Pi X$ and its normal-pass hard-assignment blocks, then draw a within-block permutation $Q$.
Let $f$ denote the SIGM core on this ordered axis, including its scans, gates, residual fusion, and normalization but excluding the entry permutation $\Pi$ and exit permutation $\Pi^{-1}$.
The normal output is $\Pi^{-1}f(x)$; the intervention returns $\Pi^{-1}Q^{-1}f(Qx)$ to the remaining network.
The normal-pass partition and $\Pi$ are reused, and $f$ neither recomputes assignments nor sorts its input again.
We repeat this intervention twenty times for each selected module.
The remaining network, including the next pruning stage, runs normally on the changed features.

\paragraph{Inverse-aligned feature and pose differences.}
The inverse-aligned feature sensitivity of the ordered-axis core $f$ is
\begin{equation}
 e_F(Q;x)=100\,
 \frac{\|Q^{-1}f(Qx)-f(x)\|_F}
 {\max\{\|f(x)\|_F,10^{-12}\}}.
 \label{eq:s_local_feature_difference}
\end{equation}
Because $\Pi^{-1}$ preserves the Frobenius norm, this equals the relative difference between the two identity-aligned module outputs.
It is a percentage of the unperturbed module output, whereas the gate-only diagnostic below normalizes by the input perturbation.
For each checkpoint, the pose decrease is the unperturbed full-test-set \mapfive minus its mean over the twenty interventions.
We average these decreases over the five paired seeds.
Within each pruning stage, \cref{tab:s_local_permutations} reports the maximum of the three module-wise mean feature sensitivities and the maximum mean pose decrease, separately for each dataset and metric.
Each column reports its own maximum over the three modules.

\begin{table}[t]
\centering
\caption{\textbf{Single-module within-cluster interventions.} I: identity; LP: fixed Gaussian low-pass; L: learned. Feature differences are inverse-aligned relative output changes (\%); pose decreases are pp. Each metric is the maximum across the three single-module means in that stage after averaging over twenty interventions and five checkpoints. Bold marks the lowest values within each stage.}
\label{tab:s_local_permutations}
\setlength{\tabcolsep}{3.5pt}
\renewcommand{\arraystretch}{1.10}
\small
\renewcommand{\tabularxcolumn}[1]{m{#1}}
\begin{tabularx}{\linewidth}{@{}cc>{\hsize=1.2\hsize\linewidth=\hsize\centering\arraybackslash}X>{\hsize=.8\hsize\linewidth=\hsize\centering\arraybackslash}X>{\hsize=1.2\hsize\linewidth=\hsize\centering\arraybackslash}X>{\hsize=.8\hsize\linewidth=\hsize\centering\arraybackslash}X@{}}
\toprule
\neutralheader{} & \neutralheader{} & \multicolumn{2}{c}{\metricheader{Feature difference (\%)}} & \multicolumn{2}{c}{\altmetricheader{Pose decrease (pp)}} \\
\cmidrule(lr){3-4}\cmidrule(lr){5-6}
\neutralheader{\multirow{-2}{*}[\dimexpr(\aboverulesep+\belowrulesep+\cmidrulewidth)/2\relax]{Stage}} & \neutralheader{\multirow{-2}{*}[\dimexpr(\aboverulesep+\belowrulesep+\cmidrulewidth)/2\relax]{Gate}} & \metricheader{YFCC100M} & \metricheader{SUN3D} & \altmetricheader{YFCC100M} & \altmetricheader{SUN3D} \\
\midrule
\tabbaseline
\cellcolor{white} & I & 4.60 & 5.40 & 0.28 & 0.20 \\
\tabstablerow
\cellcolor{white} & \tabstable{LP} & \tabstable{0.50} & \tabstable{0.62} & \tabstable{0.02} & \tabstable{0.01} \\
\tabbaseline
\cellcolor{white}\multirow{-3}{*}{1} & \oursname{L} & 0.75 & 0.95 & 0.04 & 0.03 \\
\midrule
\rowcolor{headerlilac!30!white}
\cellcolor{white} & I & 5.10 & 6.00 & 0.24 & 0.18 \\
\tabstablerow
\cellcolor{white} & \tabstable{LP} & \tabstable{0.55} & \tabstable{0.68} & \tabstable{0.01} & \tabstable{0.01} \\
\rowcolor{headerlilac!30!white}
\cellcolor{white}\multirow{-3}{*}{2} & \oursname{L} & 0.80 & 1.00 & 0.03 & 0.03 \\
\bottomrule
\end{tabularx}
\end{table}

\begin{table*}[t]
\centering
\caption{Full-input permutations on unknown splits (five checkpoints, 20 permutations each). First-stage probability MAE, retained-ID Jaccard overlaps $J_1,J_2$ (\%), and pose drops $\Delta A$, permutation variation $\sigma_{\rm perm}$, and worst drop (pp) are defined in \cref{sec:s_local_permutations}. Predictions are aligned by correspondence ID. Bold marks the best stability within each dataset.}
\label{tab:input_permutation}
\label{tab:input_permutations}
\footnotesize
\setlength{\tabcolsep}{6pt}
\renewcommand{\arraystretch}{1.12}
\begin{tabular}{@{}cccccccc@{}}
\toprule
\neutralheader{Dataset} & \neutralheader{Gates} & \metricheader{Probability MAE} & \auxheader{$J_1$} & \auxheader{$J_2$} & \altmetricheader{$\Delta A$ (pp)} & \altmetricheader{$\sigma_{\rm perm}$ (pp)} & \altmetricheader{Worst (pp)} \\
\midrule
\tabbaseline \cellcolor{white} & Identity & .0120 & 97.3 & 94.6 & $.62\tabunc{.10}$ & .24 & 1.10 \\
\tabstablerow \cellcolor{white} & \textcolor{secondc}{Fixed low-pass} & \tabstable{.0016} & \tabstable{99.7} & \tabstable{99.2} & $\tabstable{.04}\tabunc{.02}$ & \tabstable{.03} & \tabstable{.12} \\
\multirow{-3}{*}{\textbf{YFCC100M}} & \oursname{Learned spectral} & .0022 & 99.5 & 98.8 & $.08\tabunc{.03}$ & .05 & .22 \\
\midrule
\tabbaseline \cellcolor{white} & Identity & .0150 & 96.9 & 93.7 & $.42\tabunc{.09}$ & .16 & .82 \\
\tabstablerow \cellcolor{white} & \textcolor{secondc}{Fixed low-pass} & \tabstable{.0020} & \tabstable{99.6} & \tabstable{99.0} & $\tabstable{.03}\tabunc{.01}$ & \tabstable{.03} & \tabstable{.10} \\
\multirow{-3}{*}{\textbf{SUN3D}} & \oursname{Learned spectral} & .0028 & 99.3 & 98.5 & $.06\tabunc{.02}$ & .04 & .17 \\
\bottomrule
\end{tabular}
\end{table*}

\subsection{Network-input permutations}
\paragraph{Relation to network-input permutations.}
With learned gates, permuting the network input changes the mean \mapfive by only 0.08 points on YFCC100M and 0.06 points on SUN3D, compared with 0.62 and 0.42 points with identity gates, and the retained sets of both stages overlap by at least 98.5\% (\cref{tab:input_permutations}).
The whole-network tests in \cref{tab:input_permutations} apply twenty random permutations to the same original correspondence set and rerun both pruning stages allowing their assignments and retained subsets to change.
They restore correspondence identities before comparing scores and retained sets.
For each image pair and permutation trial, probability MAE is $N^{-1}\sum_{i=1}^{N}|p_i^0-\widetilde p_i|$, where both probability vectors are the first-stage scores for all original $N$ candidates before top-$K$ selection and $\widetilde p$ is aligned to their original IDs. MAE is on the probability scale.
Let $S_k^0$ and $\widetilde S_k$ be the retained original-ID sets after pruning stage $k$, with sizes $N/2$ and $N/4$ for $k=1,2$; second-stage IDs refer to the original candidate set.
The reported overlaps in percent are $J_k=100\,|S_k^0\cap\widetilde S_k|/|S_k^0\cup\widetilde S_k|$ for $k=1,2$.
We compute MAE and each $J_k$ per pair and trial, average equally over pairs and the twenty trials within each checkpoint, and then equally over the five checkpoints.
All three measures are taken before the final epipolar verification.
Let $A_{s,t}$ be the full-test-set pose \mapfive for training seed $s$ and permutation trial $t$, and let $A_s^0$ be its unperturbed value.
The paired change for a seed is $\Delta A_s=A_s^0-20^{-1}\sum_t A_{s,t}$; the reported mean and standard deviation are over the five values $\Delta A_s$.
The within-checkpoint dispersion is reported separately as $\sigma_{\mathrm{perm}}=[5^{-1}\sum_s\operatorname{Var}_t(A_{s,t})]^{1/2}$.
The worst decrease is $\max_{s,t}(A_s^0-A_{s,t})$.
Measuring scores and retained sets before verification captures every change inside the network.
Relative to identity gates, both spectral gate types reduce every sensitivity measure several-fold, and the learned gates trade slightly more variation than fixed low-pass filtering for higher accuracy.

\section{Order and gate diagnostics of the trained models}
\label{sec:s_conditional_diagnostics}

\subsection{Cluster occupancy and residual spectrum}
\begin{figure}[t]
\centering
\includegraphics[width=\linewidth]{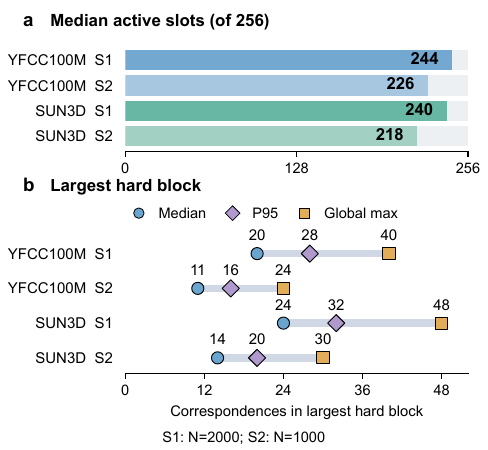}
\caption{\textbf{Hard-cluster occupancy.} Top: median nonempty slots out of $M=256$. Bottom: median, P95, and global maximum of the largest hard block $L_{\max}$ across evaluated inputs; the rightmost marker is the global maximum. Stages 1 and 2 take $N=2000$ and $1000$ correspondences, respectively.}
\label{fig:s_cluster_occupancy}
\label{tab:s_cluster_occupancy}
\end{figure}

\paragraph{Cluster occupancy.}
The results in \cref{fig:s_cluster_occupancy} describe the hard blocks obtained by assigning each correspondence to its strongest cluster slot.
Active-cluster counts are medians over evaluated image pairs and the five checkpoints, with each stage assignment counted once; quantiles and maxima refer to the per-evaluation $L_{\max}$ values.
Block sizes count hard assignments, and the fixed number $M=256$ includes empty slots.
The first and second pruning stages scan $N=2000$ and $N=1000$ correspondences, respectively.

\paragraph{Residual spectrum and conditional bounds.}
For each evaluated feature array $x$, we replace its entries within every hard block by that block's channel-wise mean to form $Hx$, and set $r=x-Hx$.
Using the unitary two-sided DFT, we compute
\begin{equation}
\begin{aligned}
 \eta_r(\omega_c)&=
 \frac{\sum_d\sum_{|\omega|\le\omega_c}|\widehat r_d(\omega)|^2}
 {\sum_d\|r_d\|_2^2},\\
 \beta(\omega_c;N,L)&=\min\!\left\{1,
 \left(\frac{\pi\omega_c L}{N}\right)^2\right\}.
\end{aligned}
 \label{eq:s_residual_diagnostic}
\end{equation}
Zero-residual inputs are reported separately.
For each dataset and stage, we use the global maximum $L$ in \cref{fig:s_cluster_occupancy} for a conservative common bound and take the maximum $\eta_r$ over the same evaluations.
In every row of \cref{tab:s_residual_bounds}, the measured low-frequency share of the residual stays an order of magnitude below the bound.

\begin{table}[t]
\centering
\caption{\textbf{Scale-separation diagnostics.} All energy fractions and bounds are percentages. Integer DFT cutoffs are used. $\eta_r^{\max}(5)$ is the maximum residual low-frequency share at cutoff five; the three $\beta$ columns are calculated from the global $L$ for the row. Bounds use these hard-block maxima.}
\label{tab:s_residual_bounds}
\setlength{\tabcolsep}{3.0pt}
\renewcommand{\arraystretch}{1.10}
\small
\renewcommand{\tabularxcolumn}[1]{m{#1}}
\begin{tabularx}{\linewidth}{@{}ccc>{\centering\arraybackslash}X>{\centering\arraybackslash}X>{\centering\arraybackslash}X>{\centering\arraybackslash}X@{}}
\toprule
\neutralheader{} & \neutralheader{} & \neutralheader{} & \auxheader{$\eta_r^{\max}(5)$} & \multicolumn{3}{c}{\altmetricheader{$\beta(\omega_c)$ (\%)}} \\
\cmidrule(lr){5-7}
\neutralheader{\multirow{-2}{*}[\dimexpr(\aboverulesep+\belowrulesep+\cmidrulewidth)/2\relax]{Dataset}} & \neutralheader{\multirow{-2}{*}[\dimexpr(\aboverulesep+\belowrulesep+\cmidrulewidth)/2\relax]{Stage}} & \neutralheader{\multirow{-2}{*}[\dimexpr(\aboverulesep+\belowrulesep+\cmidrulewidth)/2\relax]{$L$}} & \auxheader{(\%)} & \altmetricheader{$\omega_c=3$} & \altmetricheader{$5$} & \altmetricheader{$10$} \\
\midrule
\tabbaseline
YFCC100M & 1 & 40 & \cellcolor{tablestablebg}\textcolor{secondc}{0.60} & \cellcolor{headerlilac!40!white}3.55 & \cellcolor{headerlilac!40!white}9.87 & \cellcolor{headerlilac!40!white}39.48 \\
\tabbaseline
YFCC100M & 2 & 24 & \cellcolor{tablestablebg}\textcolor{secondc}{0.85} & \cellcolor{headerlilac!40!white}5.12 & \cellcolor{headerlilac!40!white}14.21 & \cellcolor{headerlilac!40!white}56.85 \\
\midrule
\rowcolor{headerlilac!30!white}
SUN3D & 1 & 48 & \cellcolor{tablestablebg}\textcolor{secondc}{0.95} & \cellcolor{headerlilac!40!white}5.12 & \cellcolor{headerlilac!40!white}14.21 & \cellcolor{headerlilac!40!white}56.85 \\
\rowcolor{headerlilac!30!white}
SUN3D & 2 & 30 & \cellcolor{tablestablebg}\textcolor{secondc}{1.40} & \cellcolor{headerlilac!40!white}7.99 & \cellcolor{headerlilac!40!white}22.21 & \cellcolor{headerlilac!40!white}88.83 \\
\bottomrule
\end{tabularx}
\end{table}

\subsection{Gate-only diagnostics}
\paragraph{Isolated-gate test at fixed positions.}
The gate-only tests use the real-valued feature array immediately after its Mamba scan and immediately before the selected gate.
For a backward gate, we use its actual reversed scan orientation and the correspondingly reversed contiguous cluster blocks.
We apply a within-block permutation $Q$ to this same array, rather than permuting an upstream Mamba input, and define
\begin{equation}
 T_G=\mathcal{F}^{-1}\operatorname{diag}(G)\mathcal{F},\qquad
 R_G=\frac{\|T_GQx-T_Gx\|_F}{\|Qx-x\|_F}.
 \label{eq:s_gate_diagnostic}
\end{equation}
The arrays in the numerator are compared at the same sequence positions.
Trials with $Qx=x$ are handled separately after checking that the output difference is numerically zero.
This normalization measures output change relative to the input perturbation, whereas \cref{eq:s_local_feature_difference} measures inverse-aligned module change relative to the original output.

\paragraph{Effective gains and uniform coefficients.}
Real-valued filtering uses rFFT/irFFT with the conjugate-symmetric effective response and real DC and Nyquist gains.
Energy sums use the equivalent unitary full DFT, or the corresponding conjugate-pair multiplicities in a one-sided implementation.
Gains are evaluated after coefficient interpolation at the actual sequence length.
For a common cutoff, we define
\begin{equation}
\begin{aligned}
 G_{\max}&=\max_{d,\omega}|G_d(\omega)|,&
 \varepsilon&=\max_{d,|\omega|>\omega_c}|G_d(\omega)|,\\
 \kappa&=\frac{\pi\omega_c L}{N},&
 b_1&=G_{\max}\kappa+\varepsilon,\\
 b_2&=\sqrt{G_{\max}^2\kappa^2+\varepsilon^2}.
\end{aligned}
\label{eq:s_gate_coefficients}
\end{equation}
The coefficient $b_1$ is the bound in \cref{eq:s_gate}, while $b_2$ follows from the quadratic form in \cref{eq:s_gate_quad} before applying $\sqrt{a^2+b^2}\le a+b$.
Both bound the fixed-position ratio $R_G$.

\begin{table}[!htbp]
\centering
\caption{\textbf{Gate-only gain and perturbation diagnostics.} For each stage and scan direction, gain extrema cover all channels, its three SIGM layers, and evaluated checkpoints; $L=48/30$ conservatively covers both unknown splits at stages 1/2. $R_G^{\max}$ is the maximum fixed-position ratio over the corresponding layers, image pairs, checkpoints, and permutation trials. Entries are dimensionless; $\varepsilon/G_{\max}$ describes relative frequency selectivity. $b_1$ is the corollary coefficient and $b_2$ its tighter quadratic-proof coefficient.}
\label{tab:s_gate_response}
\setlength{\tabcolsep}{2pt}
\renewcommand{\arraystretch}{1.10}
\small
\renewcommand{\tabularxcolumn}[1]{m{#1}}
\begin{tabularx}{\linewidth}{@{}cccccccc>{\centering\arraybackslash}X@{}}
\toprule
\neutralheader{Stage} & \neutralheader{Direction} & \neutralheader{$\omega_c$} & \metricheader{$G_{\max}$} & \metricheader{$\varepsilon$} & \metricheader{$\varepsilon/G_{\max}$} & \altmetricheader{$b_1$} & \altmetricheader{$b_2$} & \auxheader{$R_G^{\max}$} \\
\midrule
\tabbaseline
1 & Forward & 5 & 1.15 & 0.40 & 0.348 & 0.834 & 0.590 & \cellcolor{tablestablebg}\textcolor{secondc}{0.37} \\
\tabbaseline
1 & Backward & 5 & 1.25 & 0.60 & 0.480 & 1.071 & 0.763 & \cellcolor{tablestablebg}\textcolor{secondc}{0.62} \\
\midrule
\rowcolor{headerlilac!30!white}
2 & Forward & 3 & 1.12 & 0.38 & 0.339 & 0.697 & 0.495 & \cellcolor{tablestablebg}\textcolor{secondc}{0.35} \\
\rowcolor{headerlilac!30!white}
2 & Backward & 3 & 1.20 & 0.50 & 0.417 & 0.839 & 0.604 & \cellcolor{tablestablebg}\textcolor{secondc}{0.49} \\
\bottomrule
\end{tabularx}
\end{table}

\paragraph{Reading the diagnostic.}
In every row of \cref{tab:s_gate_response}, the measured ratio $R_G^{\max}$ stays below the bound $b_2$ and at most $0.62$, so each trained gate passes less than two thirds of a reordering inside the clusters.
The single-module and whole-network tests of \cref{sec:s_local_permutations} complement this fixed-position view.

\end{document}